\documentclass{article}

\usepackage{arxiv}

\usepackage[utf8]{inputenc} 
\usepackage[T1]{fontenc}    
\usepackage{hyperref}       
\usepackage{url}            
\usepackage{booktabs}       
\usepackage{amsfonts}       
\usepackage{nicefrac}       
\usepackage{microtype}      
\usepackage{lipsum}		
\usepackage{graphicx}
\usepackage{natbib}
\usepackage{doi}
\usepackage[table]{xcolor}
\graphicspath{{media/}}     
\usepackage{xcolor}
\usepackage{amssymb}
\usepackage{multirow}
\usepackage{array}
\usepackage{caption}
\definecolor{darkgreen}{rgb}{0.0, 0.5, 0.0}
\definecolor{darkred}{rgb}{0.5, 0.0, 0.0}
\usepackage[normalem]{ulem}
\useunder{\uline}{\ul}{}
\usepackage[most]{tcolorbox}
\usepackage{enumitem}
\setlist[itemize]{leftmargin=2em}

\usepackage{subcaption}
\usepackage{svg}
\svgsetup{inkscapeversion=1}
\definecolor{darkgray}{gray}{0.40}
\newcommand{\na}{\textcolor{darkgray}{--}}
\usepackage{colortbl}
\definecolor{tabhead}{RGB}{240,240,240}
\usepackage{needspace}
\usepackage{wrapfig}

\ifXeTeX
  \usepackage{newunicodechar}
  \newunicodechar{—}{\textemdash}
  \newunicodechar{’}{\textquoteright}
  \newunicodechar{→}{\ensuremath{\rightarrow}}
\fi

\definecolor{linkblue}{RGB}{30,100,200}
\definecolor{citegreen}{RGB}{0,128,96}
\definecolor{urlgray}{RGB}{80,80,80}

\makeatletter
\hypersetup{%
    colorlinks=true,
    citecolor=citegreen,   
    linkcolor=linkblue,    
    urlcolor=urlgray       
}%

\title{ReTeach: Building a Self-Teacher through Multi-Round Reflection and Retry}

\author{
      Yafeng Tang$^{1,*}$\\
    \texttt{tangyafeng8@gmail.com}
    \And
    Hao Li$^{1,*}$\\
    \texttt{erichaoli@tencent.com}
    \And
    Hongsheng Yu$^{1,\dagger}$\\
    \texttt{yannickyu@tencent.com}
    \AND
    Qiang Fu$^{1}$\\
    \texttt{leonfu@tencent.com}
}

\date{}

\renewcommand{\undertitle}{}
\renewcommand{\shorttitle}{}

\hypersetup{
pdftitle={A template for the arxiv style},
pdfsubject={},
pdfauthor={Yafeng Tang},
pdfkeywords={First keyword, Second keyword, More},
}

\begin{document}
\maketitle

\begingroup
\renewcommand{\thefootnote}{}
\footnotetext{%
    $^{*}$~Equal contribution \quad
    $^{\dagger}$~Corresponding authors\quad
    $^{1}$~Tencent Artificial Intelligence Platform Department\quad
    This work was done during Yafeng Tang's Project Up internship at Tencent.
}
\endgroup

\begin{abstract}
Self-distillation can improve reasoning without a separately trained, more capable teacher, but its effectiveness depends on how the self-teacher gains an advantage over the student.
Conditioning the teacher on reference answers or solutions can provide such an advantage, but this information may be unavailable.
Reflection offers a way to derive explicit error diagnoses and revision guidance from self-generated attempts, yet existing reflection-based methods often combine it with reference information, rich task feedback, or persistent memory.
We introduce \textbf{ReTeach}, a \textbf{Re}flective self-distillation framework that constructs its self-\textbf{Teach}er through multi-round reflection and retry using only self-generated attempts and outcome-level verification.
Starting from an unsuccessful student rollout, the teacher alternates explicit reflection with renewed attempts until success or the retry budget is exhausted, without reference answers or solutions, external diagnostic feedback, or cross-example memory.
Each failed retry informs subsequent reflection, while successful correction provides outcome-level evidence for the potential utility of the resulting teacher context.
An outcome-aware selection and weighting strategy distinguishes initially correct, reflection-corrected, and unresolved examples, assigning separate weights to their category-normalized distillation losses.
Through on-policy distillation, the student matches the teacher's context-conditioned token-level predictive distributions at prefixes of its own rollouts, transferring the benefits of iterative correction while retaining single-pass inference.
Across six benchmarks spanning mathematical reasoning, science question answering, and tool use, ReTeach improves average accuracy over GRPO by 1.39 percentage points.
Among all compared methods, including those using privileged information, ReTeach achieves the highest average accuracy on five benchmarks and ranks second on the remaining one.
These results demonstrate the effectiveness of constructing reflective self-teachers from self-generated experience and outcome-level verification to improve single-pass reasoning.
\end{abstract}


\section{Introduction}
Reinforcement learning with verifiable rewards (RLVR) has become the default recipe for post-training large language models, underpinning the strongest open reasoning models on mathematics and code, where a verifier returns a scalar reward without human annotation~\citep{lambert2025tulu,guo2025deepseekr1}. However, a binary reward per response provides no information on which intermediate steps drove the result. This sparse supervision makes training expensive: the model must discover good reasoning by comparing and contrasting many self-generated responses.
Recently, on-policy distillation (OPD) addresses this by replacing the binary reward with per-token targets from a teacher that scores the student's own rollouts~\citep{opd}. 
However, the effectiveness of OPD depends on the availability of a sufficiently capable teacher: the quality of its supervision ultimately constrains how far the student can improve, limiting the ability to advance the frontier of reasoning. This raises a natural question: can a model bootstrap its own teacher signal without relying on a separately trained, more capable teacher?

\begin{figure}[t]
    \centering
    \includegraphics[width=\linewidth,trim={0 6mm 0 5mm}]{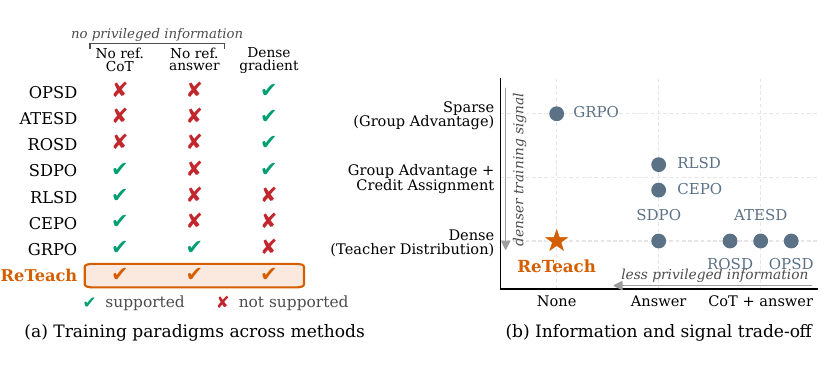}
    \caption{Comparison across existing self-improvement methods. Our ReTeach realizes denser supervision without any privileged information from original datasets. }
    \label{tab: comparison of OPSD}
\end{figure}
\vspace{-1mm}

Self-distillation derives the teacher from the model itself, but its effectiveness depends on how the teacher gains an advantage over the student. OPSD conditions its teacher on reference CoT~\citep{opsd}, while ATESD adaptively limits this exposure~\citep{han2026adaptive}. RLSD combines ground-truth answer conditioning with group-based rollout exploration~\citep{yang2026self}, whereas SDPO constructs its teacher from successful rollouts in binary-reward settings~\citep{hubotter2026reinforcement}.

Reference answers or solutions may be unavailable, while raw rollouts do not necessarily explain how to correct the student's errors.
Reflection can turn this experience into explicit error diagnoses and revision guidance~\citep{madaan2023self,shinn2023reflexion}.
Existing approaches combine reflection with task feedback, persistent memory or ground-truth answers.
RESD~\citep{zhang2026learningraresuccessrich} uses a persistent playbook and cached successful solutions with additional ground-truth answers in teacher prompts.
ReflectionCoder~\citep{ren-etal-2025-reflectioncoder} and HERO~\citep{liu2026hero} use code-execution feedback and environment observations, respectively.
We study whether effective self-teachers can instead be constructed from a model's own attempts and outcome-level verification, without reference answers or solutions, external diagnostic feedback, or cross-example memory.

We introduce \textbf{ReTeach}, a self-distillation framework that couples \emph{multi-round reflection and retry} with \emph{outcome-aware distillation}. Starting from an unsuccessful student rollout, the self-teacher explicitly reflects on the history of failed attempts and generates a new response, repeating until success or a fixed retry budget is exhausted. Retrying serves two purposes: a failed attempt provides additional evidence for subsequent reflection, while a successful attempt provides outcome-level evidence that the constructed context can support a correct response. This process uses outcome-level verification without providing the teacher with ground-truth answers, reference solutions, or externally supplied diagnostic feedback.

ReTeach uses OPD to match teacher distributions at prefixes of the original student rollout, rather than directly training on successful retries. Initially correct, reflection-corrected, and unresolved rollouts receive separately weighted, category-normalized losses. Successful correction serves as an outcome-based indicator of potential supervision utility, while unresolved examples can be downweighted or excluded. The student retains single-pass inference. Figure~\ref{tab: comparison of OPSD} compares ReTeach with existing self-improvement methods. Our contributions are as follows:
\begin{itemize}[leftmargin=*]
    \item We introduce a multi-round reflection-and-retry process for self-teacher construction. Rather than stopping at a retrospective diagnosis, ReTeach generates renewed attempts whose failures inform subsequent reflection, developing teacher contexts through iterative correction without access to reference answers or solutions.

    \item We propose outcome-aware selection and category-level weighting that connect correction outcomes to distillation supervision. By distinguishing initially correct, reflection-corrected, and unresolved rollouts, ReTeach balances preserving existing behavior with learning from reflective contexts, while accounting for the less certain utility of unresolved examples.

    \item Across mathematical reasoning, science question answering, and tool-use tasks, ReTeach achieves average accuracies of 65.62\%, 66.71\%, and 62.72\%, respectively, consistently outperforming GRPO (64.32\%, 65.43\%, and 60.85\%). Among the compared methods, including those that use external privileged information, ReTeach achieves the highest average accuracy on five of the six benchmarks and ranks second on the remaining one.
\end{itemize}

\section{Background}
In this section, we introduce several key concepts that form the foundation of self-distillation, the core of ReTeach. Then, we compare existing self-distillation methods by how the privileged advantage is exploited in Figure~\ref{tab: comparison of OPSD}, with a detailed survey of each baseline deferred to Section~\ref{sec: related work}.

\textbf{Step-Level Chain-of-Thought Reasoning.}
Chain-of-thought (CoT) prompting and training encourage LLMs to generate intermediate reasoning steps before committing to a final answer~\citep{wei2022chain,kojima2022large}, which decomposes a hard problem into tractable sub-steps and substantially improves accuracy on mathematical and multi-step reasoning tasks ~\citep{yu2024metamath}. Beyond accuracy, the explicit trace makes the model's computation inspectable and allows supervision to be attached to individual steps rather than only to the outcome, offering denser feedback and improved interpretability ~\citep{uesato2022solving,lightman2024lets}. CoT enables step-level supervision through process reward models or token-level proxies, but requires auxiliary models or extra annotation ~\citep{lightman2024lets,wang2024mathshepherd,cui2026process, zhang2025generative}, which motivates researchers to obtain dense targets from distillation instead.

\textbf{Token-Level Knowledge Distillation.}
Knowledge distillation transfers a teacher $\pi_{\mathrm{tea}}$ into a student $\pi_\theta$ by matching next-token distributions instead of one-hot labels, providing a dense target at every position~\citep{hinton2015distilling,kim2016sequence}. Offline imitation of teacher traces~\citep{yu2024metamath,guo2025deepseekr1} suffers from a train--inference distribution mismatch and compounding exposure bias~\citep{agarwal2024onpolicy,gu2024minillm}. On-policy distillation (OPD) methods construct targets from samples generated by the current policy or its improved variants, thereby converting reinforcement signals into supervised updates~\citep{agarwal2024onpolicy,gu2024minillm,opd}. The matching is performed on the student's own rollouts,
\begin{equation}
\mathcal{L}_{\mathrm{OPD}}(\theta)=\mathbb{E}_{y\sim\pi_\theta(\cdot\mid x)}\Big[\textstyle\sum_{t}\mathrm{KL}\big(\pi_\theta(\cdot\mid x,y_{<t})\,\|\,\pi_{\mathrm{tea}}(\cdot\mid x,y_{<t})\big)\Big],
\label{eq:opd}
\end{equation}
so the mismatch disappears, and the teacher--student log-ratio acts as a per-token advantage ~\citep{agarwal2024onpolicy,hubotter2026reinforcement}. Still, OPD needs a stronger external teacher. Such a teacher is computationally expensive to obtain and upper-bounds the student's capability~\citep{opd,wang2026denser}.

\paragraph{On-Policy Self-Distillation.}Recently, on-policy self-distillation (OPSD)~\citep{opsd} has emerged as a popular approach that eliminates the need for a separate teacher by running a single model under two information conditions ~\citep{hubotter2026reinforcement}. The student observes only the prompt, while the teacher also sees the privileged context $c$, available exclusively during training. The teacher is then the same policy conditioned on $c$,
\begin{equation}
\pi_{\mathrm{tea}}(\cdot\mid x,y_{<t}):=\mathrm{sg}\big(\pi_\theta(\cdot\mid x,c,y_{<t})\big),
\label{eq:selfteacher}
\end{equation}
where $\mathrm{sg}(\cdot)$ denotes stop-gradient and keeps the teacher from co-evolving with the student. Since $c$ is withheld during inference, the teacher remains better informed than the student ~\citep{hubotter2026reinforcement,yang2026self}, and distilling it back provides dense token-level supervision at no additional model cost. While methods represented by OPSD have demonstrated strong empirical performance, they still rely on some form of external privileged information. We instead seek to remove this dependence, allowing the model to correct itself through iterative self-reflection.



\section{Methodology}

\begin{figure}[t]
    \centering
        \includegraphics[width=\linewidth,trim={0 25mm 0 0}]{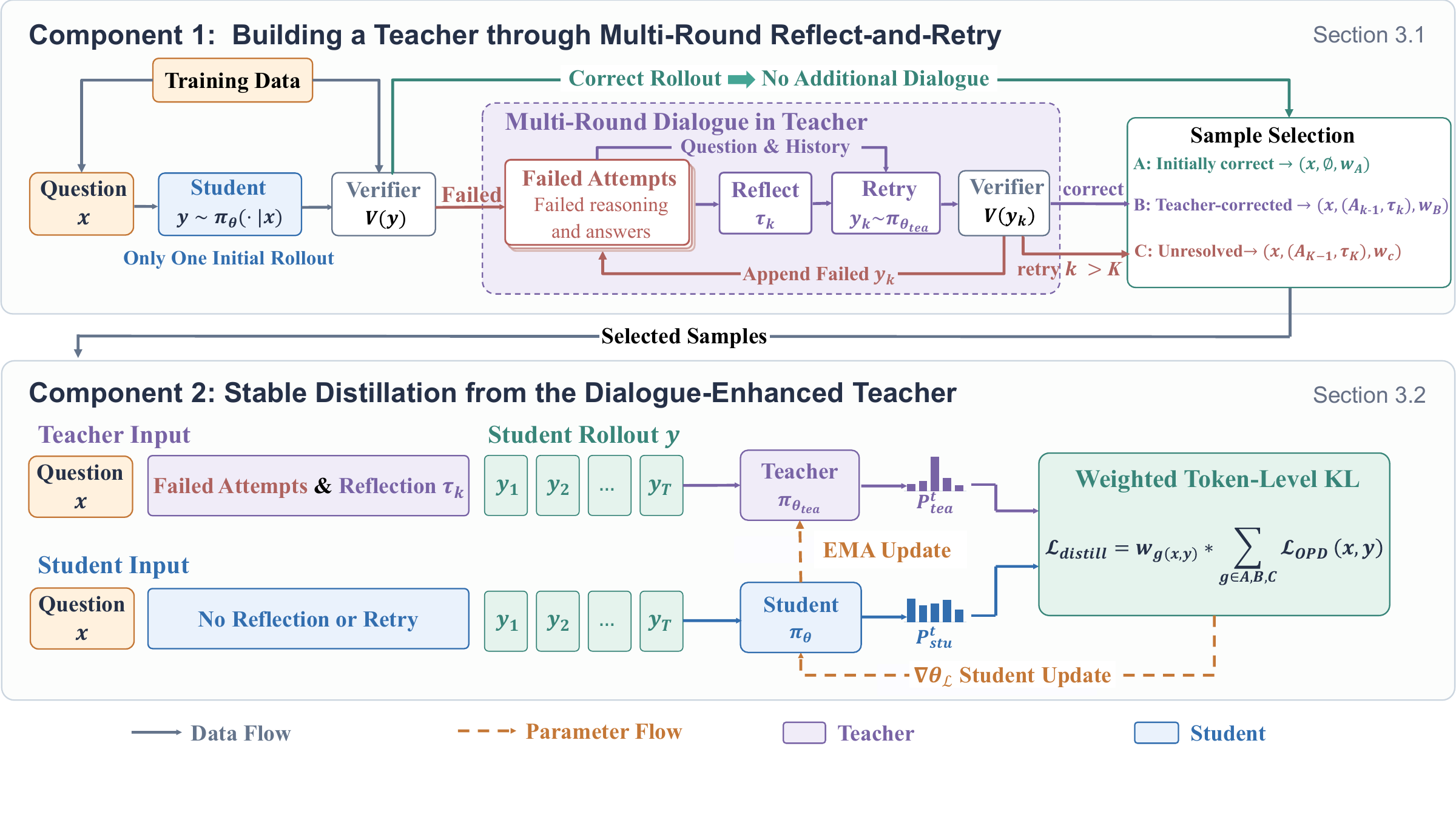}
    \caption{An overview of our ReTeach method. }
    \label{fig:method_overview}
\end{figure}

In this section, we propose the ReTeach algorithm, which builds a teacher through multi-round reflect-and-retry for self-distillation as shown in Figure~\ref{fig:method_overview}.

\subsection{Building a Teacher through Multi-Round Reflect-and-Retry}
\label{sec:reflection}

Inspired by recent work on iterative reasoning and self-refinement~\citep{radha2024iterationthoughtleveraginginner,madaan2023self}, \textbf{ReTeach} constructs a deliberative teacher through multi-round interaction. The central idea is that a model can recover from an initially unsuccessful response by repeatedly revisiting its previous attempts, explicitly reflecting on potential errors, and trying again.

For each question $x$, the student first samples a response $y \sim \pi_\theta(\cdot \mid x)$, which is evaluated by a verifier $r=\mathcal{V}(y)\in\{0,1\}$. If $r=1$, no refinement is needed, and we set $c(x)=\emptyset$. Otherwise, the teacher initiates a multi-round \emph{reflect-and-retry} process from the failed student response.
Specifically, let
\begin{equation}
A_{k-1} = (y,y_1,\ldots,y_{k-1})
\end{equation}
denote the attempt history available at round $k$, where $y$ is the initial student response and $y_i$ is the teacher's retry at round $i$. At each round, the teacher first generates an explicit reflection $\tau_k$ based on the question and all previous attempts, and then produces a new response $y_k$ conditioned additionally on the current reflection:
\begin{equation}
\begin{aligned}
\tau_k
&\sim
\pi_{\theta_{\mathrm{tea}}}
\bigl(\cdot \mid x,A_{k-1}\bigr),\\
y_k
&\sim
\pi_{\theta_{\mathrm{tea}}}
\bigl(\cdot \mid x,A_{k-1},\tau_k\bigr).
\end{aligned}
\label{eq:dialogue_refinement}
\end{equation}
Importantly, $A_{k-1}$ contains only previous attempts; reflections from earlier rounds are not carried forward. Each reflection therefore serves as a transient intermediate step that guides the retry within the current round. The teacher is instructed to enclose $\tau_k$ within \texttt{<reflect>} and \texttt{</reflect>} tags before generating $y_k$, making error reflection an explicit part of the refinement process. The complete prompt template is provided in Appendix~C.
After each round, the verifier evaluates the retry $y_k$. The reflect-and-retry process terminates when a retry $y_k$ satisfies $\mathcal{V}(y_k)=1$, or after $K$ unsuccessful rounds, where $K$ is the maximum number of rounds permitted by the context budget. Let $T$ denote the final round. We define the teacher content as
\begin{equation}
c(x)=\bigl(A_{T-1},\tau_T\bigr)
     =\bigl(y,y_1,\ldots,y_{T-1},\tau_T\bigr),
\label{eq:teacher_content}
\end{equation}
which consists of all attempts preceding the final retry together with the final-round reflection. In this way, the teacher constructs its refinement context entirely from self-generated attempts, without access to externally supplied reference solutions or corrective feedback.

\subsection{Stable On-Policy Distillation from the Deliberative Teacher}
\label{sec: ema}
We transfer the capabilities of the deliberative teacher to the single-turn student through on-policy distillation. 
For each question $x$, distillation is performed on the student rollout $y\sim\pi_\theta(\cdot\mid x)$.
The teacher and student are evaluated on the same rollout $y$, but with different contexts: the student observes only the question, whereas the teacher additionally receives the deliberative content $c(x)$ constructed in Section~\ref{sec:reflection}. 
Specifically, at each token position $t$ of student rollout $y$,
\begin{equation}
\begin{aligned}
p_{\mathrm{stu}}^t
&=
\pi_\theta(\cdot\mid x,y_{<t}),\\
p_{\mathrm{tea}}^t
&=
\mathrm{sg}\!\left[
\pi_{\theta_{\mathrm{tea}}}
(\cdot\mid x,c(x),y_{<t})
\right],
\end{aligned}
\label{eq:teacher_student_policies}
\end{equation}
where $\mathrm{sg}[\cdot]$ stops gradients through the teacher. Since
$y$ is sampled from the current student policy, the distillation
objective remains on-policy. 
The asymmetric contexts allow the student to internalize the benefits of reflect-and-retry.

We minimize the token-level Jensen--Shannon divergence between the two predictive distributions:
\begin{equation}
\mathcal{L}_{\mathrm{OPD}}(x,y)
=
\frac{1}{|y|}
\sum_{t=1}^{|y|}
D_{\mathrm{JS}}
\left(p_{\mathrm{tea}}^t,p_{\mathrm{stu}}^t\right),
\label{eq:opd_loss}
\end{equation}
where
\begin{equation}
D_{\mathrm{JS}}(p,q) = \frac{1}{2}D_{\mathrm{KL}}(p\|m) + \frac{1}{2}D_{\mathrm{KL}}(q\|m), \qquad m= 1/2 p+ 1/2 q.
\label{eq:js_divergence}
\end{equation}

\paragraph{EMA Teacher.}
The teacher must evolve with the student while providing stable
distillation targets. Directly tracking it with the student can
lead to drastic fluctuations during optimization, whereas keeping it
frozen prevents it from benefiting from subsequent improvements.
We therefore initialize the teacher from the student and update its
parameters after each student optimization step using an exponential
moving average:
\begin{equation}
\theta_{\mathrm{tea}}
\leftarrow
\lambda\theta_{\mathrm{tea}}
+
(1-\lambda)\theta,
\label{eq:ema_update}
\end{equation}
where $0<\lambda<1$ controls the degree of temporal smoothing. This
update allows the teacher to track improvements in the student while
maintaining stable targets for distillation.

\paragraph{Outcome-Aware Sample Weighting.}
We divide the on-policy rollouts into three categories according to their initial correctness and the outcome of reflection and retry:
\begin{itemize}[leftmargin=*]
    \item Category A: Initially correct rollouts, for which $\mathcal{V}(y)=1$ and $c(x)=\emptyset$. The teacher and student receive the same prompt and token prefix, so their predictive distributions differ only through their parameters. Distillation from the EMA teacher regularizes the student's predictions against a temporally lagged reference, helping preserve previously correct behavior.
    \item Category B: Initially incorrect rollouts that are corrected through reflection and retry, for which $\mathcal{V}(y)=0$ and $\mathcal{V}(y_T)=1$. As described in Section~\ref{sec:reflection}, the teacher conditions on $x$, the history $A_{T-1}$, and the final reflection $\tau_T$, giving $c(x)=(A_{T-1},\tau_T)$. Its token-level predictive distributions are evaluated along the original student rollout $y$, rather than the corrected attempt $y_T$. The successful attempt $y_T$ provides outcome-level evidence that this context $c(x)$ can support a correct response, motivating its use for corrective distillation.
    \item Category C: Initially incorrect rollouts that remain unresolved after reflection and retry. Teacher distributions are computed in the same way as for Category B, but the resulting context is not associated with a verified successful attempt. Although such contexts may contain useful partial reasoning, their reliability as distillation supervision is less certain than that of Category B.
\end{itemize}
Let $\mathcal{D}_A$, $\mathcal{D}_B$, and $\mathcal{D}_C$ denote the three category-specific subsets of the current training batch. We define the category-level OPD loss as
\begin{equation}
\mathcal{L}_g
=
\begin{cases}
\displaystyle
\frac{1}{|\mathcal{D}_g|}
\sum_{(x,y)\in\mathcal{D}_g}
\mathcal{L}_{\mathrm{OPD}}(x,y),
& |\mathcal{D}_g|>0,\\[6pt]
0, & |\mathcal{D}_g|=0,
\end{cases}
\qquad g\in\{A,B,C\}.
\label{eq:category_opd}
\end{equation}
The overall distillation objective is
\begin{equation}
\mathcal{L}_{\mathrm{distill}}
=
w_A\mathcal{L}_A+w_B\mathcal{L}_B+w_C\mathcal{L}_C,
\label{eq:weighted_opd}
\end{equation}
where $w_A$, $w_B$, and $w_C$ are the prescribed category weights. An absent category contributes neither loss nor gradient, and the weights of the remaining categories are not renormalized. 
For each nonempty category, averaging normalizes its loss by the category size, while its coefficient in the overall objective remains $w_g$.


\section{Experiment}
\subsection{Experiment setup}
\paragraph{Models and datasets.}\label{sec: models and datasets}

We evaluate the within-dataset held-out performance in science question answering and tool use, and evaluate the cross-dataset performance in mathematical reasoning:
\begin{itemize}[leftmargin=*]
    \item \textbf{Within-Dataset Held-Out Performance.} For science question answering, we use the undergraduate-level reasoning subsets (L3) of SciKnowEval~\citep{sciknoweval} covering Physics and Materials. For tool-use, we use ToolAlpaca~\citep{toolalpaca}, where the model is given a tool API specification and a natural-language user request and is required to generate the corresponding structured tool call. We follow the same split construction and evaluation configurations recommended in~\cite{hubotter2026reinforcement}. Dataset sources and acquisition details are in Appendix~\ref{experiment_details}.
    \item \textbf{Cross-Dataset Performance.} For mathematical reasoning, we follow OPSD~\citep{opsd} and train on up to 30,000 problem--solution
    pairs sampled from the mathematical reasoning subset of OpenThoughts~\citep{guha2025openthoughtsdatarecipesreasoning}. Each training example contains a mathematical problem and a step-by-step solution. We evaluate the trained model on external competition-level benchmarks, including AIME 2024, AIME 2025, and the February HMMT 2025~\citep{dekoninck2026matharena}. Dataset construction and acquisition details are in Appendix~\ref{experiment_details}.
\end{itemize}

\paragraph{Baselines.}
We compare ReTeach with four post-training baselines, covering the standard RLVR method and the most widely compared on-policy self-distillation methods to date, with CEPO being the most recent (May 2026). We also report the initial checkpoint as a non-post-trained reference:
\begin{itemize}[leftmargin=*]
    \item \textbf{GRPO}~\citep{grpo} is a standard reinforcement learning
    with verifiable rewards (RLVR) method. It computes group-relative
    advantages from sequence-level verifier rewards and applies them in a
    token-level policy optimization objective. 

    \item \textbf{OPSD}~\citep{opsd} is an on-policy self-distillation
    method. It aligns the student policy with a positive teacher
    distribution associated with verifier-accepted responses using a
    per-token KL-divergence objective.

    \item \textbf{RLSD}~\citep{yang2026self} uses a positive teacher
    associated with verifier-accepted responses to construct a
    stop-gradient likelihood ratio, which rescales the magnitude of the
    token-level signal.

    \item \textbf{CEPO}~\citep{heakl2026cepo} builds a contrastive token-level
  weight from a positive teacher on verifier-accepted responses and a negative
  teacher on verifier-rejected rollouts

    \item \textbf{RESD}~\citep{zhang2026learningraresuccessrich} builds the teacher context from environment feedback verified against the dataset ground truth, by conditioning the self-teacher on a self-generated reflection over each failure, a persistent playbook of curated lessons, and cached successful rollouts.
\end{itemize}

\paragraph{Implementation Details.} We conduct all experiments initialized from Qwen3-8B~\citep{yang2025qwen3technicalreport}, using LoRA~\citep{hu2022iclr-lora} on 8~H20 GPUs. To keep a fair and sound comparison, we follow the training and evaluation protocols widely adopted in prior work~\citep{opsd, hubotter2026reinforcement}, covering the decoding configuration and the reported metrics. We list the full set of hyperparameters in Table~\ref{tab: training Parameter} and Table~\ref{tab: evaluation Parameter}. Full details are provided in Appendix~\ref{experiment_details}.


\subsection{Main Results}
Table~\ref{tab: main_results} summarizes performance under within-dataset and cross-dataset evaluation.
ReTeach ranks first on five of the six benchmarks and second on SciKnowEval-Material, achieving the highest overall average of $65.50\%$, exceeding GRPO and the strongest baseline overall, CEPO, by $1.39$ and $1.35$ percentage points, respectively.
Relative to CEPO, ReTeach also has a lower seed-level standard deviation on most datasets.
Within datasets, ReTeach performs best on Physics and ToolAlpaca, improving over GRPO by $2.28$ and $1.87$ percentage points, respectively, while CEPO leads on Material.
For cross-dataset evaluation, ReTeach achieves the highest avg@12 on all three math benchmarks, with a mean accuracy of $65.62\%$, compared with $64.32\%$ for GRPO and $62.93\%$ for CEPO.
Together, these results demonstrate competitive within-dataset performance and strong cross-dataset generalization without supplying reference reasoning traces or ground-truth answers to the teacher.

\begin{table}[t]
\centering
\captionsetup{position=top,skip=8pt}
\caption{Performance comparison across cross-dataset and within-dataset evaluation. We report the mean$_{\pm\text{std}}$ over three random seeds. The cross-dataset block reports Avg@12 on three competition-level math benchmarks under the same metric recommended in~\citep{opsd}. The within-dataset block reports Avg@16 on the science and tool-use tasks under the configuration of~\citep{hubotter2026reinforcement}.
\textbf{Bold} marks the best mean in each row, and \uline{underline} marks the second best. The within-dataset results of OPSD are marked as $\na$, because these datasets provide no reference CoT; its Total Average is therefore the mean over the three math benchmarks only.}
\label{tab: main_results}
\resizebox{\linewidth}{!}{%
\begin{tabular}{@{}clccccccc@{}}
\specialrule{\heavyrulewidth}{\abovetopsep}{0pt}
\rowcolor{tabhead}
\multicolumn{2}{c}{\cellcolor{tabhead}\textbf{Method}} & \textbf{Qwen3-8B} & \textbf{GRPO} & \textbf{OPSD} & \textbf{RLSD} & \textbf{RESD} & \textbf{CEPO} & \textbf{ReTeach} \\
\specialrule{\lightrulewidth}{0pt}{\belowrulesep}
\multirow{3}{*}{\rotatebox[origin=c]{90}{\textbf{Within}}} & SciKnowEval-Physics & 59.84$_{\pm 0.32}$ & 61.96$_{\pm 1.25}$ & $\na$ & 61.90$_{\pm 0.83}$ & 61.09$_{\pm 0.54}$ & \uline{63.39}$_{\pm 1.45}$ & \textbf{64.24}$_{\pm 1.05}$ \\
        & SciKnowEval-Material & 64.89$_{\pm 1.50}$ & 68.91$_{\pm 2.03}$ & $\na$ & 68.35$_{\pm 0.39}$ & 68.02$_{\pm 0.97}$ & \textbf{70.83}$_{\pm 1.35}$ & \uline{69.17}$_{\pm 0.97}$ \\
        & ToolUse-ToolAlpaca & 57.90$_{\pm 0.21}$ & 60.85$_{\pm 0.30}$ & $\na$ & 60.08$_{\pm 0.77}$ & 60.94$_{\pm 0.37}$ & \uline{61.86}$_{\pm 1.62}$ & \textbf{62.72}$_{\pm 0.20}$ \\
        \midrule
\multirow{3}{*}{\rotatebox[origin=c]{90}{\textbf{Cross}}} & Math-AIME24 & 76.94$_{\pm 0.57}$ & 76.48$_{\pm 0.47}$ & \uline{76.95}$_{\pm 0.61}$ & 76.94$_{\pm 0.99}$ & 75.83$_{\pm 0.27}$ & 76.11$_{\pm 1.64}$ & \textbf{78.61}$_{\pm 0.99}$ \\
        & Math-AIME25 & 66.94$_{\pm 0.67}$ & \uline{70.37}$_{\pm 0.69}$ & 67.67$_{\pm 2.21}$ & 69.72$_{\pm 0.68}$ & 66.11$_{\pm 0.70}$ & 68.52$_{\pm 0.34}$ & \textbf{70.46}$_{\pm 1.02}$ \\
        & Math-HMMT25 & 44.72$_{\pm 0.41}$ & 46.11$_{\pm 0.60}$ & \uline{46.67}$_{\pm 0.22}$ & 46.39$_{\pm 0.68}$ & 44.35$_{\pm 0.37}$ & 44.17$_{\pm 1.71}$ & \textbf{47.78}$_{\pm 1.42}$ \\
        \midrule
Total & Average & 61.87 & 64.11 & 63.76 & 63.90 & 62.72 & \uline{64.15} & \textbf{65.50} \\ \bottomrule
\end{tabular}%
}
\end{table}

\subsection{Ablation Studies and Discussions}\label{sec: Ablation}

\begin{figure}[t]
    \centering
    \includegraphics[width=\linewidth,trim={0 1em 0 0}]{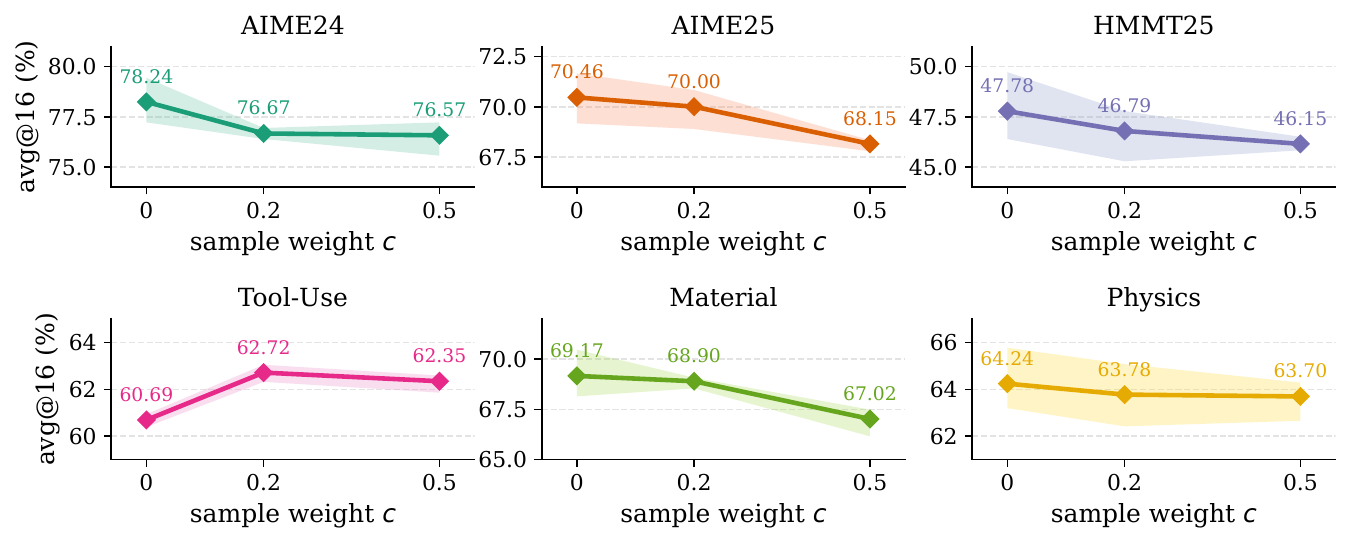}
    \caption{Ablation on the weight $w_C$ of unresolved rollouts for ReTeach on Qwen3-8B, with $w_A=w_B=0.5$ fixed. Lines and shaded bands show the mean and standard deviation over three seeds.}
    \label{fig: sample weight}
\end{figure}

\paragraph{Effect of Weighting Unresolved Rollouts.}\label{sec: wc-ablation}
Our outcome-aware weighting uses whether the teacher ultimately solves a problem through reflection and retry as a proxy for the usefulness of the resulting distillation signal.
We therefore examine whether rollouts that remain unresolved (Category C) should contribute to distillation despite the absence of a verified successful outcome.
To assess their contribution, we fix $w_A=w_B=0.5$ and vary $w_C\in\{0,0.2,0.5\}$, where $w_C=0$ excludes unresolved rollouts from the distillation objective and $w_C=0.5$ assigns equal weights to the three category-level mean losses.
Figure~\ref{fig: sample weight} reports performance across three training seeds on the three Math benchmarks, Tool-Use, and the two SciKnowEval datasets.
On the Math and SciKnowEval benchmarks, including unresolved rollouts provides no consistent improvement, and increasing $w_C$ to $0.5$ further reduces performance.
In contrast, Tool-Use benefits from $w_C=0.2$ but deteriorates at $w_C=0.5$.
In the inspected Tool-Use rollouts, most predicted function calls have required arguments that match the reference, but differ in optional argument values and are therefore rejected by the exact-match verifier.
These cases suggest that unresolved contexts can retain useful partial information despite failing outcome verification, which may explain the benefit of assigning them a modest weight.
However, potentially useful information in Category C is mixed with errors and noise, so increasing $w_C$ may amplify unreliable supervision and degrade student performance.
Overall, the results support task-dependent weighting of unresolved rollouts rather than uniformly treating their supervision as either useful or unusable.

\begin{wrapfigure}[16]{r}{0.48\linewidth}
\vspace{-4pt}
\centering
\captionsetup{font=small,skip=4pt}
\includegraphics[width=\linewidth,trim={5pt 5pt 0 0},clip]{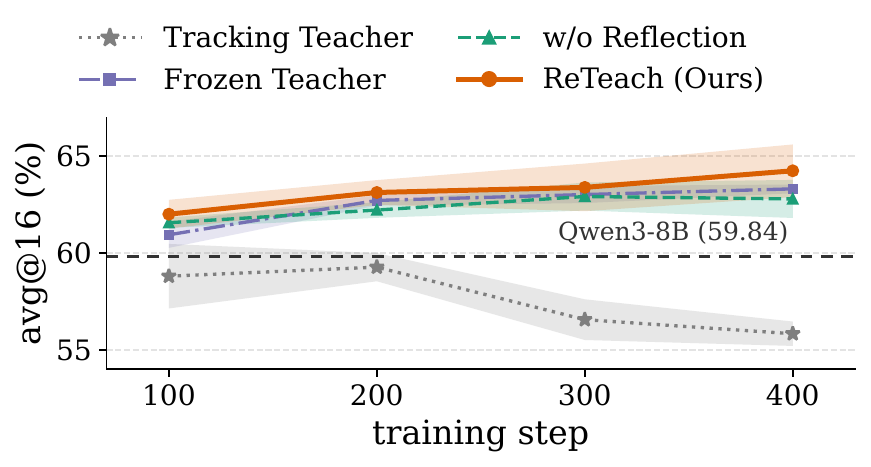}
\caption{Ablation of EMA and reflection designs on SciKnowEval-Physics (avg@16). Lines and bands show the mean and standard deviation over three seeds; the dashed line marks the base model Qwen3-8B.}
\label{fig: ema reflect ablation}
\end{wrapfigure}

\paragraph{Effectiveness of Explicit Reflection.}
At each reflection-and-retry round, the teacher generates an explicit reflection $\tau_k$ between \texttt{<reflect>} and \texttt{</reflect>}, followed by a new attempt $y_k$ (Eq.~\eqref{eq:dialogue_refinement}). We ablate this reflection step while retaining iterative retries: at round $k$, the teacher samples directly from the unsuccessful-attempt history $A_{k-1}$:
\begin{equation}
y_k \sim \pi_{\theta_{\mathrm{tea}}} \bigl(\cdot \mid x,A_{k-1}\bigr). 
\label{eq:reflection_ablation}
\end{equation}
The teacher context used for distillation is correspondingly reduced to $c(x)=A_{T-1}$, without the final reflection $\tau_T$. As shown by \textit{w/o Reflection} in Figure~\ref{fig: ema reflect ablation}, removing explicit reflection reduces student accuracy from $64.24\%$ to $62.91\%$, a drop of $1.33$ percentage points. This result supports the contribution of explicit reflection to ReTeach's self-distillation process and is consistent with prior findings that explicit reflection can serve as feedback to improve model performance across successive attempts~\citep{madaan2023self, shinn2023reflexion}.

\begin{figure}[htbp]
    \centering
    \includegraphics[width=\linewidth,clip]{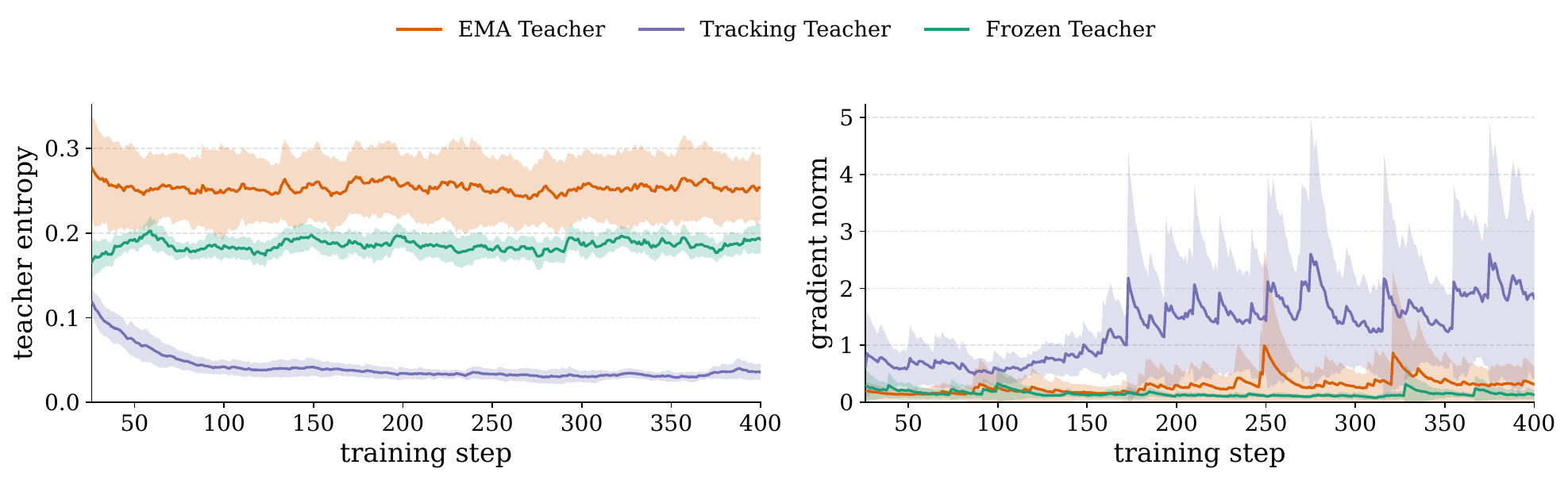}
    \caption{Teacher entropy (left) and gradient norm (right) on SciKnowEval-Physics during training under tracking, frozen, and EMA teacher updates. Lines and bands show the mean and standard deviation over three seeds.}
    \label{fig: entropy and grad}
\end{figure}

\paragraph{Effectiveness of EMA Target Network.} 
The EMA teacher stabilizes distillation targets while incorporating student improvements. To isolate its effect, we retain multi-round reflection-and-retry and compare it with a tracking teacher synchronized with the student and a teacher frozen at the base model (Figure~\ref{fig: ema reflect ablation}). (1) \uline{The tracking teacher undermines the training stability.} With a tracking teacher, the student fails to outperform the base model and gradually declines as training progressed.
Two mechanisms help explain this result. First, direct synchronization transfers the student's parameter changes to the teacher without temporal smoothing, causing the distillation targets to change rapidly. Second, on initially correct rollouts, the teacher and student share the same prompt and token prefix, so minimizing their JSD regularizes the student's deviation from the teacher's predictions.
This regularization gradient can appear only as the student diverges from the teacher and vanishes when their parameters and predictive distributions coincide. Figure~\ref{fig: entropy and grad} further characterizes this failure mode through teacher entropy and gradient norm. The tracking teacher's entropy falls from $0.129$ to $0.036$ nats, while its gradient norm rises from $0.29$ to $2.31$ and peaks at $25.27$, an order of magnitude above the other configurations. With the teacher refreshed from the student at every update, the two distributions nearly coincide, and token-level divergence concentrates on positions where the student is already confident. The resulting large but uninformative gradient reinforces existing predictions rather than correcting them, consistent with the observed entropy collapse.
(2) \uline{Freezing stabilizes the targets but prevents teacher adaptation.} The student trained with a frozen teacher consistently outperforms the base model but trails ReTeach. 
Because the teacher's parameters remain fixed, it cannot incorporate the capabilities acquired by the student to improve its own reasoning.
Its entropy remains near $0.18$ nats and its gradient norm near $0.19$ (Figure~\ref{fig: entropy and grad}), indicating stable but unadaptive supervision.
The EMA teacher provides a middle ground by absorbing student improvements through a lagged update, which keeps distillation targets from changing too quickly and preserves a delayed reference on initially correct rollouts. It has the lowest and most stable gradient norm ($0.08$--$0.14$) and the highest entropy, increasing from $0.21$ to $0.24$ nats. These dynamics support the role of EMA in preserving target stability while allowing the teacher to track student improvements.

\begin{figure}[ht]
  \centering
  \vspace{-1em}
  \begin{subfigure}[t]{0.48\linewidth}
    \centering
    \includegraphics[width=\linewidth,trim={2mm 2mm 0 3mm},clip]{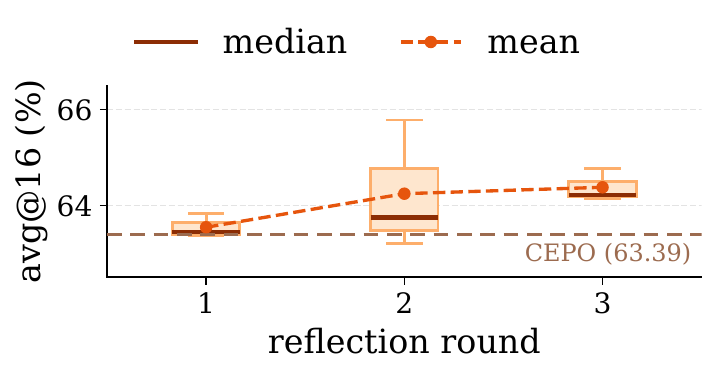}
    \caption{Ablation study on reflection rounds.}
    \label{tab: round}
  \end{subfigure}
  \hfill
  \begin{subfigure}[t]{0.5\linewidth}
    \centering
    \includegraphics[width=\linewidth,clip]{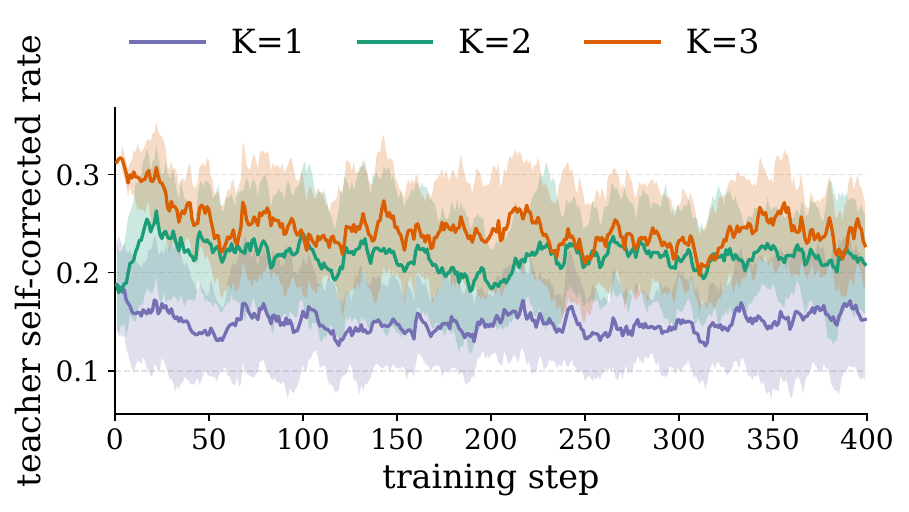}
    \caption{Teacher self-corrected rate with different rounds $K$.}
    \label{fig: rate_of_B}
  \end{subfigure}
  \caption{Reflection-round analysis for ReTeach on SciKnowEval-Physics.}
  \label{fig: round_analysis}
\end{figure}\vspace{-1em}

\paragraph{Effect of the Maximum Number of Rounds.}
Increasing the maximum number of reflection-and-retry rounds $K$ gives the teacher more opportunities to correct unsuccessful attempts and potentially solve harder problems. To examine how this translates into student performance, we evaluate ReTeach on SciKnowEval-Physics with $K\in\{1,2,3\}$. As shown in Figure~\ref{tab: round}, the corresponding avg@16 accuracies are $63.54\%$, $64.24\%$, and $64.38\%$, respectively. Increasing $K$ from $1$ to $2$ yields a gain of $0.70$ percentage points, whereas increasing it from $2$ to $3$ adds only $0.14$ percentage points, although the variance in final accuracy across training seeds decreases markedly. These results indicate diminishing returns in mean accuracy, with an additional benefit in cross-seed consistency at $K=3$. 
To explain this diminishing return, we track during training the fraction of Category B examples that the teacher succeeds in correcting through reflection and retry (Figure~\ref{fig: rate_of_B}). Raising $K$ from $1$ to $2$ increases this fraction substantially, whereas raising it from $2$ to $3$ increases it only marginally. 
Because additional rounds change the teacher's supervision only for examples that are eventually corrected, the marginal gain in student accuracy closely follows the marginal increase in this fraction: once a further round recovers few additional corrected examples, it supplies little new guidance beyond what earlier rounds already provide. The third round nevertheless reduces the variability of the final accuracy across training seeds, so its remaining benefit lies more in consistency than in mean accuracy. Although $K=3$ attains the highest mean accuracy among the tested settings, it also substantially increases teacher inference time during training. We therefore use $K=2$ as the default to balance performance gains against training cost.

\section{Related Work}
\label{sec: related work}

\paragraph{Self-Teacher Construction.}
Self-teachers provide dense distillation targets and refine token-level credit assignment in reinforcement learning.
Following context distillation~\citep{snell2022learning}, many methods construct stronger self-teachers by conditioning them on information unavailable to the student.
OPSD~\citep{opsd} conditions its teacher on reference solutions, while RLSD~\citep{yang2026self} and GEAR~\citep{li2026geargranularityadaptiveadvantagereweighting} use ground-truth answers.
Extensions adapt reference granularity~\citep{han2026adaptive} or contrast teachers conditioned on correct and incorrect answers~\citep{heakl2026cepo,pan2026rlcsdreinforcementlearningcontrastive}.
Others exploit feedback or model-generated experience: SDPO~\citep{hubotter2026reinforcement} uses runtime feedback or successful rollouts from the same group; SC-GRPO~\citep{shan2026learningsolutionsselfconditionedcredit} and CAST~\citep{li2026castnonprivilegedclippedasymmetric} use verified self-generated rollouts or trajectory correctness rather than reference answers.

\paragraph{Reflection-Based Self-Improvement.}
Self-Refine~\citep{madaan2023self} and Reflexion~\citep{shinn2023reflexion} revisit earlier attempts through inference-time feedback, refinement, and reflection, whereas Re-ReST~\citep{dou2025rerestreflectionreinforcedselftraininglanguage} repairs trajectories through environment-based reflection for off-policy self-training.
Beyond trajectory repair, RICL~\citep{chen2026retrospectiveincontextlearningtemporal} generates retrospective step-level feedback and uses feedback-conditioned changes in action probabilities for credit assignment, while ROSD~\citep{zhao2026rosd} uses reflection to identify error locations and restrict distillation to the corresponding suffixes.
ReflectionCoder~\citep{ren-etal-2025-reflectioncoder} distills final-code predictions conditioned on offline, execution-guided multi-round reflection sequences.
Its data construction includes execution-based filtering of unresolved and initially successful attempts.
RESD~\citep{zhang2026learningraresuccessrich} combines retrospective reflection with ground-truth answer feedback, a persistent playbook, and cached successful solutions; HERO~\citep{liu2026hero} derives turn-level reflective hints from completed interactions and environment observations.
ReTeach instead constructs teacher contexts online through multi-round reflection and retry, using only self-generated attempts and outcome-level verification, without reference answers or solutions, external diagnostic feedback, or cross-example memory.
Failed retries provide new evidence for subsequent reflection, while successful correction supplies outcome-level evidence for the potential utility of the resulting teacher context.
Whereas RESD rebalances successful and failed rollouts, ReTeach retains initially correct examples and further separates initially failed rollouts into reflection-corrected and unresolved categories, using correction outcomes to weight category-normalized distillation losses.

\section{Conclusion}
\label{sec:conclusion}

In this work, we introduced ReTeach, a self-distillation framework that constructs a stronger teacher through reasoning-process asymmetry. Instead of relying on externally supplied privileged information, the teacher revisits failed student responses through multiple rounds of explicit reflection and retry. The resulting deliberative context enables the teacher to provide informative token-level targets on rollouts sampled from the current student policy. The outcome-aware classwise weighting balances the learning signals from initially correct, successfully refined, and unresolved problems, while an EMA update maintains a stable yet evolving teacher.

Experiments on mathematical reasoning, science question answering, and tool-use benchmarks demonstrate the consistent effectiveness of ReTeach. Even compared with methods that leverage privileged information, such as ground-truth answers or reference solutions, ReTeach achieves the best performance on five of the six benchmarks and ranks second on the remaining one. These results show that the advantage of a self-constructed teacher need not stem solely from information asymmetry: altering how the same model reasons can itself yield an effective source of supervision. More broadly, our findings suggest that capabilities elicited through iterative deliberation can be internalized into a more efficient single-pass policy, offering a promising direction for improving reasoning models when only outcome-level verification is available.


\newpage

\bibliography{iclr2027_conference}

@inproceedings{
opsd,
title={Self-Distilled Reasoner: On-Policy Self-Distillation for Large Language Models},
author={Siyan Zhao and Zhihui Xie and Mengchen Liu and Jing Huang and Guan Pang and Feiyu Chen and Aditya Grover},
booktitle={Forty-third International Conference on Machine Learning},
year={2026},
url={https://openreview.net/forum?id=Jpxfof0EaS}
}

@inproceedings{hubotter2026reinforcement,
  title = {Reinforcement Learning via Self-Distillation},
  author = {Hübotter, Jonas and Lübeck, Frederike and Behric, Lejs and Baumann, Anton and Bagatella, Marco and Marta, Daniel and Hakimi, Ido and Shenfeld, Idan and Kleine Buening, Thomas and Guestrin, Carlos and Krause, Andreas},
  year = {2026},
  booktitle = {International Conference on Machine Learning (2026)},
}

@inproceedings{hu2022iclr-lora,
  title     = {{LoRA: Low-Rank Adaptation of Large Language Models}},
  author    = {Hu, Edward J and Shen, Yelong and Wallis, Phillip and Allen-Zhu, Zeyuan and Li, Yuanzhi and Wang, Shean and Wang, Lu and Chen, Weizhu},
  booktitle = {International Conference on Learning Representations},
  year      = {2022},
  url       = {https://mlanthology.org/iclr/2022/hu2022iclr-lora/}
}

@misc{yang2025qwen3technicalreport,
      title={Qwen3 Technical Report}, 
      author={An Yang and Anfeng Li and Baosong Yang and Beichen Zhang and Binyuan Hui and Bo Zheng and Bowen Yu and Chang Gao and Chengen Huang and Chenxu Lv and Chujie Zheng and Dayiheng Liu and Fan Zhou and Fei Huang and Feng Hu and Hao Ge and Haoran Wei and Huan Lin and Jialong Tang and Jian Yang and Jianhong Tu and Jianwei Zhang and Jianxin Yang and Jiaxi Yang and Jing Zhou and Jingren Zhou and Junyang Lin and Kai Dang and Keqin Bao and Kexin Yang and Le Yu and Lianghao Deng and Mei Li and Mingfeng Xue and Mingze Li and Pei Zhang and Peng Wang and Qin Zhu and Rui Men and Ruize Gao and Shixuan Liu and Shuang Luo and Tianhao Li and Tianyi Tang and Wenbiao Yin and Xingzhang Ren and Xinyu Wang and Xinyu Zhang and Xuancheng Ren and Yang Fan and Yang Su and Yichang Zhang and Yinger Zhang and Yu Wan and Yuqiong Liu and Zekun Wang and Zeyu Cui and Zhenru Zhang and Zhipeng Zhou and Zihan Qiu},
      year={2025},
      eprint={2505.09388},
      archivePrefix={arXiv},
      primaryClass={cs.CL},
      url={https://arxiv.org/abs/2505.09388}, 
}

@misc{yang2026self,
  title         = {Self-Distilled RLVR},
  author        = {Yang, Chenxu and Qin, Chuanyu and Si, Qingyi and Chen, Minghui and Gu, Naibin and Yao, Dingyu and Lin, Zheng and Wang, Weiping and Wang, Jiaqi and Duan, Nan},
  year          = {2026},
  eprint        = {2604.03128},
  archivePrefix = {arXiv},
  url           = {https://arxiv.org/abs/2604.03128}
}

@misc{heakl2026cepo,
  title         = {CEPO: RLVR Self-Distillation using Contrastive Evidence Policy Optimization},
  author        = {Heakl, Ahmed and Shaker, Abdelrahman M. and Mohamed, Youssef and Elbadry, Rania and Fetouh, Omar and Khan, Fahad Shahbaz and Khan, Salman},
  year          = {2026},
  eprint        = {2605.19436},
  archivePrefix = {arXiv},
  url           = {https://arxiv.org/abs/2605.19436}
}

@misc{han2026adaptive,
  title         = {Adaptive Teacher Exposure for Self-Distillation in LLM Reasoning},
  author        = {Han, Zihao and Zhang, Tiangang and Wang, Huaibin and Sun, Yilun},
  year          = {2026},
  eprint        = {2605.11458},
  archivePrefix = {arXiv},
  url           = {https://arxiv.org/abs/2605.11458}
}

@misc{zhao2026rosd,
  title         = {ROSD: Reflective On-Policy Self-Distillation for Language Model Reasoning across Domains},
  author        = {Zhao, Ziqi and Ma, Xinyu and Yang, Liu and Feng, Yujie and Shi, Daiting and He, Jingzhou and Xin, Xin and Ren, Zhaochun and Wu, Xiao-Ming},
  year          = {2026},
  eprint        = {2605.28014},
  archivePrefix = {arXiv},
  url           = {https://arxiv.org/abs/2605.28014}
}

@misc{wang2026denser,
  title         = {Denser $\neq$ Better: Limits of On-Policy Self-Distillation for Continual Post-Training},
  author        = {Wang, Meng and Zhao, Haohan and Liu, Wenzhuo and Yang, Lu and Liu, Geng and Guo, Haiyang and Xie, Guo-Sen and Meng, Gaofeng and Liu, Hongbin and Zhu, Fei},
  year          = {2026},
  eprint        = {2607.01763},
  archivePrefix = {arXiv},
  url           = {https://arxiv.org/abs/2607.01763}
}

@article{opd,
  author = {Kevin Lu and Thinking Machines Lab},
  title = {On-Policy Distillation},
  journal = {Thinking Machines Lab: Connectionism},
  year = {2025},
  note = {https://thinkingmachines.ai/blog/on-policy-distillation},
  doi = {10.64434/tml.20251026},
}

@inproceedings{
agarwal2024onpolicy,
title={On-Policy Distillation of Language Models: Learning from Self-Generated Mistakes},
author={Rishabh Agarwal and Nino Vieillard and Yongchao Zhou and Piotr Stanczyk and Sabela Ramos Garea and Matthieu Geist and Olivier Bachem},
booktitle={The Twelfth International Conference on Learning Representations},
year={2024},
url={https://openreview.net/forum?id=3zKtaqxLhW}
}

@inproceedings{
shinn2023reflexion,
title={Reflexion: language agents with verbal reinforcement learning},
author={Noah Shinn and Federico Cassano and Ashwin Gopinath and Karthik R Narasimhan and Shunyu Yao},
booktitle={Thirty-seventh Conference on Neural Information Processing Systems},
year={2023},
url={https://openreview.net/forum?id=vAElhFcKW6}
}

@misc{grpo,
      title={DeepSeekMath: Pushing the Limits of Mathematical Reasoning in Open Language Models}, 
      author={Zhihong Shao and Peiyi Wang and Qihao Zhu and Runxin Xu and Junxiao Song and Xiao Bi and Haowei Zhang and Mingchuan Zhang and Y. K. Li and Y. Wu and Daya Guo},
      year={2024},
      eprint={2402.03300},
      archivePrefix={arXiv},
      primaryClass={cs.CL},
      url={https://arxiv.org/abs/2402.03300}, 
}

@inproceedings{
guha2025openthoughtsdatarecipesreasoning,
title={OpenThoughts: Data Recipes for Reasoning Models},
author={Etash Kumar Guha and Ryan Marten and Sedrick Keh and Negin Raoof and Georgios Smyrnis and Hritik Bansal and Marianna Nezhurina and Jean Mercat and Trung Vu and Zayne Rea Sprague and Ashima Suvarna and Benjamin Feuer and Leon Liangyu Chen and Zaid Khan and Eric Frankel and Sachin Grover and Caroline Choi and Niklas Muennighoff and Shiye Su and Wanjia Zhao and John Yang and Shreyas Pimpalgaonkar and Kartik sharma and Charlie Cheng-Jie Ji and Yichuan Deng and Sarah M Pratt and Vivek Ramanujan and Jon Saad-Falcon and Stutee Acharya and Jeffrey Li and Achal Dave and Alon Albalak and Kushal Arora and Blake Wulfe and Chinmay Hegde and Greg Durrett and Sewoong Oh and Mohit Bansal and Saadia Gabriel and Aditya Grover and Kai-Wei Chang and Vaishaal Shankar and Aaron Gokaslan and Mike A Merrill and Tatsunori Hashimoto and Yejin Choi and Jenia Jitsev and Reinhard Heckel and Maheswaran Sathiamoorthy and Alex Dimakis and Ludwig Schmidt},
booktitle={The Fourteenth International Conference on Learning Representations},
year={2026},
url={https://openreview.net/forum?id=7xjoTuaNmN}
}

@misc{sciknoweval,
  title         = {SciKnowEval: Evaluating Multi-level Scientific Knowledge of Large Language Models},
  author        = {Feng, Kehua and Ding, Keyan and Wang, Weijie and Zhuang, Xiang and Wang, Zeyuan and Qin, Ming and Zhao, Yu and Yao, Jianhua and Zhang, Qiang and Chen, Huajun},
  year          = {2024},
  eprint        = {2406.09098},
  archivePrefix = {arXiv},
  url           = {https://arxiv.org/abs/2406.09098}
}

@misc{toolalpaca,
  title         = {ToolAlpaca: Generalized Tool Learning for Language Models with 3000 Simulated Cases},
  author        = {Tang, Qiaoyu and Deng, Ziliang and Lin, Hongyu and Han, Xianpei and Liang, Qiao and Sun, Le},
  year          = {2023},
  eprint        = {2306.05301},
  archivePrefix = {arXiv},
  url           = {https://arxiv.org/abs/2306.05301}
}

@inproceedings{
dekoninck2026matharena,
title={Beyond Benchmarks: MathArena as an Evaluation Platform for Mathematics with {LLM}s},
author={Jasper Dekoninck and Nikola Jovanovi{\'c} and Tim Gehrunger and K{\'a}ri R{\"o}gnvaldsson and Ivo Petrov and Chenhao Sun and Martin Vechev},
booktitle={3rd AI for Math Workshop: Toward Self-Evolving Scientific Agents},
year={2026},
url={https://openreview.net/forum?id=DmPE4byHuN}
}

@article{madaan2023self,
  title={Self-refine: Iterative refinement with self-feedback},
  author={Madaan, Aman and Tandon, Niket and Gupta, Prakhar and Hallinan, Skyler and Gao, Luyu and Wiegreffe, Sarah and Alon, Uri and Dziri, Nouha and Prabhumoye, Shrimai and Yang, Yiming and others},
  journal={Advances in neural information processing systems},
  volume={36},
  pages={46534--46594},
  year={2023}
}

@misc{radha2024iterationthoughtleveraginginner,
      title={Iteration of Thought: Leveraging Inner Dialogue for Autonomous Large Language Model Reasoning},
      author={Santosh Kumar Radha and Yasamin Nouri Jelyani and Ara Ghukasyan and Oktay Goktas},
      year={2024},
      eprint={2409.12618},
      archivePrefix={arXiv},
      primaryClass={cs.CL},
      url={https://arxiv.org/abs/2409.12618},
}

@inproceedings{
wei2022chain,
title={Chain of Thought Prompting Elicits Reasoning in Large Language Models},
author={Jason Wei and Xuezhi Wang and Dale Schuurmans and Maarten Bosma and brian ichter and Fei Xia and Ed H. Chi and Quoc V Le and Denny Zhou},
booktitle={Advances in Neural Information Processing Systems},
editor={Alice H. Oh and Alekh Agarwal and Danielle Belgrave and Kyunghyun Cho},
year={2022},
url={https://openreview.net/forum?id=_VjQlMeSB_J}
}

@inproceedings{kojima2022large,
author = {Kojima, Takeshi and Gu, Shixiang Shane and Reid, Machel and Matsuo, Yutaka and Iwasawa, Yusuke},
title = {Large language models are zero-shot reasoners},
year = {2022},
isbn = {9781713871088},
publisher = {Curran Associates Inc.},
address = {Red Hook, NY, USA},
booktitle = {Proceedings of the 36th International Conference on Neural Information Processing Systems},
articleno = {1613},
numpages = {15},
location = {New Orleans, LA, USA},
series = {NIPS '22}
}

@inproceedings{
yu2024metamath,
title={MetaMath: Bootstrap Your Own Mathematical Questions for Large Language Models},
author={Longhui Yu and Weisen Jiang and Han Shi and Jincheng YU and Zhengying Liu and Yu Zhang and James Kwok and Zhenguo Li and Adrian Weller and Weiyang Liu},
booktitle={The Twelfth International Conference on Learning Representations},
year={2024},
url={https://openreview.net/forum?id=N8N0hgNDRt}
}

@misc{uesato2022solving,
  title         = {Solving Math Word Problems with Process- and Outcome-Based Feedback},
  author        = {Uesato, Jonathan and Kushman, Nate and Kumar, Ramana and Song, H. Francis and Siegel, Noah Y. and Wang, Lisa and Creswell, Antonia and Irving, Geoffrey and Higgins, Irina},
  year          = {2022},
  eprint        = {2211.14275},
  archivePrefix = {arXiv},
  url           = {https://arxiv.org/abs/2211.14275}
}

@inproceedings{
lightman2024lets,
title={Let's Verify Step by Step},
author={Hunter Lightman and Vineet Kosaraju and Yuri Burda and Harrison Edwards and Bowen Baker and Teddy Lee and Jan Leike and John Schulman and Ilya Sutskever and Karl Cobbe},
booktitle={The Twelfth International Conference on Learning Representations},
year={2024},
url={https://openreview.net/forum?id=v8L0pN6EOi}
}

@inproceedings{wang2024mathshepherd,
    title = "Math-Shepherd: Verify and Reinforce {LLM}s Step-by-step without Human Annotations",
    author = "Wang, Peiyi  and
      Li, Lei  and
      Shao, Zhihong  and
      Xu, Runxin  and
      Dai, Damai  and
      Li, Yifei  and
      Chen, Deli  and
      Wu, Yu  and
      Sui, Zhifang",
    editor = "Ku, Lun-Wei  and
      Martins, Andre  and
      Srikumar, Vivek",
    booktitle = "Proceedings of the 62nd Annual Meeting of the Association for Computational Linguistics (Volume 1: Long Papers)",
    month = aug,
    year = "2024",
    address = "Bangkok, Thailand",
    publisher = "Association for Computational Linguistics",
    url = "https://aclanthology.org/2024.acl-long.510/",
    doi = "10.18653/v1/2024.acl-long.510",
    pages = "9426--9439",
}

@article{
cui2026process,
title={Process Reinforcement through Implicit Rewards},
author={Ganqu Cui and Lifan Yuan and Zefan Wang and Hanbin Wang and Yuchen Zhang and Jiacheng Chen and Wendi Li and Bingxiang He and Yuchen Fan and Tianyu Yu and Qixin Xu and Weize Chen and Jiarui Yuan and Huayu Chen and Kaiyan Zhang and Xingtai Lv and Shuo Wang and Yuan Yao and Xu Han and Hao Peng and Yu Cheng and Zhiyuan Liu and Maosong Sun and Bowen Zhou and Ning Ding},
journal={Transactions on Machine Learning Research},
issn={2835-8856},
year={2026},
url={https://openreview.net/forum?id=9SkkifLopZ},
note={}
}

@inproceedings{
zhang2025generative,
title={Generative Verifiers: Reward Modeling as Next-Token Prediction},
author={Lunjun Zhang and Arian Hosseini and Hritik Bansal and Mehran Kazemi and Aviral Kumar and Rishabh Agarwal},
booktitle={The Thirteenth International Conference on Learning Representations},
year={2025},
url={https://openreview.net/forum?id=Ccwp4tFEtE}
}

@misc{hinton2015distilling,
  title         = {Distilling the Knowledge in a Neural Network},
  author        = {Hinton, Geoffrey and Vinyals, Oriol and Dean, Jeff},
  year          = {2015},
  eprint        = {1503.02531},
  archivePrefix = {arXiv},
  url           = {https://arxiv.org/abs/1503.02531}
}

@inproceedings{kim2016sequence,
    title = "Sequence-Level Knowledge Distillation",
    author = "Kim, Yoon  and
      Rush, Alexander M.",
    editor = "Su, Jian  and
      Duh, Kevin  and
      Carreras, Xavier",
    booktitle = "Proceedings of the 2016 Conference on Empirical Methods in Natural Language Processing",
    month = nov,
    year = "2016",
    address = "Austin, Texas",
    publisher = "Association for Computational Linguistics",
    url = "https://aclanthology.org/D16-1139/",
    doi = "10.18653/v1/D16-1139",
    pages = "1317--1327"
}

@misc{guo2025deepseekr1,
      title={DeepSeek-R1: Incentivizing Reasoning Capability in LLMs via Reinforcement Learning}, 
      author={DeepSeek-AI},
      year={2025},
      eprint={2501.12948},
      archivePrefix={arXiv},
      primaryClass={cs.CL},
      url={https://arxiv.org/abs/2501.12948}, 
}

@inproceedings{
gu2024minillm,
title={Mini{LLM}: Knowledge Distillation of Large Language Models},
author={Yuxian Gu and Li Dong and Furu Wei and Minlie Huang},
booktitle={The Twelfth International Conference on Learning Representations},
year={2024},
url={https://openreview.net/forum?id=5h0qf7IBZZ}
}

@misc{snell2022learning,
      title={Learning by Distilling Context},
      author={Charlie Snell and Dan Klein and Ruiqi Zhong},
      year={2022},
      eprint={2209.15189},
      archivePrefix={arXiv},
      primaryClass={cs.CL},
      url={https://arxiv.org/abs/2209.15189},
}

@inproceedings{
lambert2025tulu,
title={Tulu 3: Pushing Frontiers in Open Language Model Post-Training},
author={Nathan Lambert and Jacob Morrison and Valentina Pyatkin and Shengyi Huang and Hamish Ivison and Faeze Brahman and Lester James Validad Miranda and Alisa Liu and Nouha Dziri and Xinxi Lyu and Yuling Gu and Saumya Malik and Victoria Graf and Jena D. Hwang and Jiangjiang Yang and Ronan Le Bras and Oyvind Tafjord and Christopher Wilhelm and Luca Soldaini and Noah A. Smith and Yizhong Wang and Pradeep Dasigi and Hannaneh Hajishirzi},
booktitle={Second Conference on Language Modeling},
year={2025},
url={https://openreview.net/forum?id=i1uGbfHHpH}
}

@misc{li2026geargranularityadaptiveadvantagereweighting,
      title={GEAR: Granularity-Adaptive Advantage Reweighting for LLM Agents via Self-Distillation},
      author={Sijia Li and Yuchen Huang and Zifan Liu and Yanping Li and Jingjing Fu and Li Zhao and Jiang Bian and Ling Zhang and Jun Zhang and Rui Wang},
      year={2026},
      eprint={2605.11853},
      archivePrefix={arXiv},
      primaryClass={cs.LG},
      url={https://arxiv.org/abs/2605.11853},
}

@misc{pan2026rlcsdreinforcementlearningcontrastive,
      title={RLCSD: Reinforcement Learning with Contrastive On-Policy Self-Distillation},
      author={Leyi Pan and Shuchang Tao and Yunpeng Zhai and Lingzhe Zhang and Zhaoyang Liu and Bolin Ding and Aiwei Liu and Lijie Wen},
      year={2026},
      eprint={2606.11709},
      archivePrefix={arXiv},
      primaryClass={cs.LG},
      url={https://arxiv.org/abs/2606.11709},
}

@misc{shan2026learningsolutionsselfconditionedcredit,
      title={Learning from Own Solutions: Self-Conditioned Credit Assignment for Reinforcement Learning with Verifiable Rewards},
      author={Yingyu Shan and Yuhang Guo and Zihao Cheng and Zeming Liu and Xiangrong Zhu and Xinyi Wang and Jiashu Yao and Wei Lin and Hongru Wang and Heyan Huang},
      year={2026},
      eprint={2606.18810},
      archivePrefix={arXiv},
      primaryClass={cs.LG},
      url={https://arxiv.org/abs/2606.18810},
}

@misc{li2026castnonprivilegedclippedasymmetric,
      title={CAST: Non-Privileged Clipped Asymmetric Self-Teaching with Advantage Flipping for GRPO},
      author={Yang Li and Gongle Xue and Yijia Guo and Yuheng Yuan and Liwen Hu and Lei Ma},
      year={2026},
      eprint={2606.00172},
      archivePrefix={arXiv},
      primaryClass={cs.AI},
      url={https://arxiv.org/abs/2606.00172},
}

@inproceedings{dou2025rerestreflectionreinforcedselftraininglanguage,
    title = "Re-{R}e{ST}: Reflection-Reinforced Self-Training for Language Agents",
    author = "Dou, Zi-Yi  and
      Yang, Cheng-Fu  and
      Wu, Xueqing  and
      Chang, Kai-Wei  and
      Peng, Nanyun",
    editor = "Al-Onaizan, Yaser  and
      Bansal, Mohit  and
      Chen, Yun-Nung",
    booktitle = "Proceedings of the 2024 Conference on Empirical Methods in Natural Language Processing",
    month = nov,
    year = "2024",
    address = "Miami, Florida, USA",
    publisher = "Association for Computational Linguistics",
    url = "https://aclanthology.org/2024.emnlp-main.861/",
    doi = "10.18653/v1/2024.emnlp-main.861",
    pages = "15394--15411",
}

@inproceedings{
chen2026retrospectiveincontextlearningtemporal,
title={Retrospective In-Context Learning for Temporal Credit Assignment with Large Language Models},
author={Wentse Chen and Jiayu Chen and Fahim Tajwar and Hao Zhu and Xintong Duan and Ruslan Salakhutdinov and Jeff Schneider},
booktitle={The Thirty-ninth Annual Conference on Neural Information Processing Systems},
year={2025},
url={https://openreview.net/forum?id=QAVpe6a3rp}
}

@misc{zhang2026learningraresuccessrich,
  title         = {Learning with Rare Success but Rich Feedback via Reflection-Enhanced Self-Distillation},
  author        = {Yuwei Zhang and Sha Li and Changlong Yu and Qin Lu and Shuowei Jin and Chengyu Dong and Haoran Liu and Ilgee Hong and Xintong Li and Zhenyu Shi and Bing Yin and Jingbo Shang},
  year          = {2026},
  eprint        = {2605.12741},
  archivePrefix = {arXiv},
  primaryClass  = {cs.LG},
  url           = {https://arxiv.org/abs/2605.12741}
}

@misc{liu2026hero,
  title = {{HERO}: Hindsight-Enhanced Reflection from Environment Observations for Agentic Self-Distillation},
  author = {Liu, Haoran and Zhang, Yuwei and Li, Xiyao and Lyu, Bohan and Shang, Jingbo},
  year = {2026},
  eprint = {2606.11559},
  archivePrefix = {arXiv},
  primaryClass = {cs.AI},
  url = {https://arxiv.org/abs/2606.11559}
}

@inproceedings{ren-etal-2025-reflectioncoder,
  author       = {Houxing Ren and
                  Mingjie Zhan and
                  Zhongyuan Wu and
                  Aojun Zhou and
                  Junting Pan and
                  Hongsheng Li},
  editor       = {Wanxiang Che and
                  Joyce Nabende and
                  Ekaterina Shutova and
                  Mohammad Taher Pilehvar},
  title        = {ReflectionCoder: Learning from Reflection Sequence for Enhanced One-off
                  Code Generation},
  booktitle    = {Proceedings of the 63rd Annual Meeting of the Association for Computational
                  Linguistics (Volume 1: Long Papers), {ACL} 2025, Vienna, Austria,
                  July 27 - August 1, 2025},
  pages        = {9999--10020},
  publisher    = {Association for Computational Linguistics},
  year         = {2025},
  url          = {https://doi.org/10.18653/v1/2025.acl-long.494},
  doi          = {10.18653/V1/2025.ACL-LONG.494},
  bibsource    = {dblp computer science bibliography, https://dblp.org}
}
\bibliographystyle{iclr2027_conference}

\newpage

\appendix
\section{Experiment details}\label{experiment_details}
\paragraph{Datasets.}
For mathematical reasoning, we follow the training recipe of OPSD~\citep{opsd}: the mathematical-reasoning subset of OpenThoughts~\citep{guha2025openthoughtsdatarecipesreasoning}, sampling up to 30{,}000 problem--solution pairs with chain-of-thought. Cross-dataset evaluation uses AIME 2024, AIME 2025, and February HMMT 2025~\citep{dekoninck2026matharena}, the same three benchmarks as OPSD. For science question answering we use the undergraduate-level (L3) Physics and Materials subsets of SciKnowEval~\citep{sciknoweval}, and for tool use we use ToolAlpaca~\citep{toolalpaca}. On these two tasks we follow the within-dataset held-out splits and evaluation protocol of SDPO~\citep{hubotter2026reinforcement}. For training-time outcome verification, we do not follow OPSD or SDPO in exposing a reference CoT or the ground-truth answer to the teacher. Like GRPO, ReTeach uses an RLVR outcome verifier $\mathcal{V}(y)\in\{0,1\}$ that only marks an attempt as correct or incorrect.

\paragraph{Category weights.}
Table~\ref{tab: training Parameter} sets $w_A=w_B=0.5$ on all tasks, $w_C=0$ on Math and Science, and $w_C=0.2$ on Tool-Use. These $w_C$ values were chosen from the ablation in Figure~\ref{fig: sample weight}, not from a separate validation split. Unresolved rollouts did not improve Math or SciKnowEval, while Tool-Use improved at $w_C=0.2$ and degraded at $w_C=0.5$. We therefore discard Category C on closed-answer tasks and keep a modest $w_C$ on Tool-Use. Detailed results and the analysis of this effect are given in Section~\ref{sec: wc-ablation}.

\paragraph{Training setup.}
We build all methods on Qwen3-8B and train on three tasks, \emph{Math}, \emph{Tool-Use}, and \emph{Science}. The training configuration is shared across the three tasks and follows OPSD~\citep{opsd}, including the thinking mode of each policy and the sampling hyperparameters. The teacher therefore keeps its thinking mode enabled throughout training while the student is trained with it disabled. For the GRPO-style baselines including GRPO, CEPO, RLSD, and RESD, we use 8 rollouts per question, and set the clipping thresholds to $\epsilon_{\text{low}} = 0.2$ and $\epsilon_{\text{high}} = 0.28$. All remaining training hyperparameters are identical to those of ReTeach and appear in Table~\ref{tab: training Parameter}.

\paragraph{Evaluation protocol.}
The three tasks are reported following the protocol of the closest published method on each dataset, which keeps every comparison against numbers produced under the same protocol. On the within-dataset evaluations, \emph{Science} and \emph{Tool-Use}, we follow SDPO~\citep{hubotter2026reinforcement} and report avg@16 with the thinking mode disabled, consistent with the setting in which the student is trained. 
On the cross-dataset \emph{Math} evaluation, we follow OPSD and report results on the three benchmarks used there, including AIME24, AIME25, and HMMT25. We evaluate with avg@12 and enable the thinking mode at inference, so that our numbers remain directly comparable to those reported by OPSD. Table~\ref{tab: evaluation Parameter} summarizes both protocols.



\begin{table}[htbp]
\centering
\caption{Training configuration for ReTeach}
\renewcommand{\arraystretch}{1.3}
\begin{tabular}{@{}lcccc@{}}
\toprule
 & \textbf{Parameter} & \textbf{Math} & \textbf{Science} & \textbf{Tool-Use} \\
\midrule

 & Sample Weight A      & 0.5    & 0.5    & 0.5    \\
 & Sample Weight B      & 0.5  & 0.5  & 0.5  \\
 & Sample Weight C      & 0    & 0    & 0.2  \\
 & student max tokens   & 8192 & 1024 & 1024 \\
 & teacher max tokens   & 8192 & 4096 & 4096 \\
 & Effective Batch Size & 32   & 32   & 32   \\
\multirow{-7}{*}{\rotatebox[origin=c]{90}{\textbf{Data}}}
 & Number of Rollouts   & 1    & 1    & 1    \\
\midrule
 & LoRA Rank     & 64  & 64  & 64  \\
 & LoRA $\alpha$ & 128 & 128 & 128 \\
\multirow{-4}{*}{\rotatebox[origin=c]{90}{\textbf{LoRA}}}
 & LoRA Target Modules
 & \multicolumn{3}{c}{\begin{tabular}[c]{@{}c@{}}
     q\_proj, k\_proj, v\_proj, o\_proj \\
     gate\_proj, up\_proj, down\_proj
   \end{tabular}} \\
\midrule

 & Inference Engine & vLLM & vLLM & vLLM \\
 & Temperature      & 1.1  & 1.1  & 1.1  \\
 & Top-$p$            & 0.95 & 0.95 & 0.95 \\
\multirow{-4}{*}{\rotatebox[origin=c]{90}{\textbf{Rollout}}}
 & Top-$K$            & 20   & 20   & 20   \\
\midrule

 & Teacher Think Mode       & Enabled  & Enabled  & Enabled  \\
 & Student Think Mode       & Disabled & Disabled & Disabled \\
 & Max Grad Norm           & 0.1      & 0.1      & 0.1      \\
 & Teacher Temperature     & 0.7      & 0.7      & 0.7      \\
  & EMA Decay $\lambda$     & 0.9999   & 0.9999   & 0.9999   \\
\multirow{-6}{*}{\rotatebox[origin=c]{90}{\textbf{Training}}}
 & Training Steps          & 400      & 400      & 400      \\
\midrule
\end{tabular}%
\label{tab: training Parameter}
\end{table}

\begin{table}[htbp]
\centering
\caption{Evaluation configuration for all methods}
\renewcommand{\arraystretch}{1.3}
\begin{tabular}{@{}lcccc@{}}
\toprule
 & \textbf{Parameter} & \textbf{Math} & \textbf{Science} & \textbf{Tool-Use} \\ \midrule
\multirow{2}{*}{\rotatebox[origin=c]{90}{\textbf{Data}}}
 & Samples per Prompt & 12    & 16    & 16    \\
 & Max New Tokens     & 38912 & 38912 & 38912 \\ \midrule
\multirow{4}{*}{\rotatebox[origin=c]{90}{\textbf{Evaluation}}}
 & Think Mode   & Enabled & Disabled & Disabled \\
 & Temperature  & 0.6     & 0.6      & 0.6      \\
 & Top-$p$        & 0.95    & 0.95     & 0.95     \\
 & Top-$k$        & -1      & -1       & -1       \\ \bottomrule
\end{tabular}\label{tab: evaluation Parameter}
\end{table}


\definecolor{boxproblem}{RGB}{245,247,250}
\definecolor{boxprompt}{RGB}{250,247,240}
\definecolor{boxgen}{RGB}{243,249,244}
\definecolor{boxrule}{RGB}{150,160,175}

\newtcolorbox{paperbox}[2]{%
  enhanced, breakable,
  title={#2},
  fonttitle=\bfseries\small,
  colback=#1, colframe=boxrule,
  coltitle=black,
  boxrule=0.5pt,
  left=5pt, right=5pt, top=4pt, bottom=4pt,
  arc=1.5pt,
  boxsep=3pt,
}

\section{Data Samples}\label{app: data samples}

We train and evaluate on four data sources: \emph{Math} (competition-level mathematics), and the \emph{Physics}, \emph{Material}, and \emph{Tool-Use} subsets used in our science and tool-use evaluation. This section shows one training question from each source, reproduced verbatim from the corresponding data file. Each question is paired with a verifiable outcome only; no reference solution or human-written chain of thought is used anywhere in training.

\begin{paperbox}{boxproblem}{Data sample: Math}
Let proposition $P$: The sets of solutions to the inequalities $a_{1}x^{2}+b_{1}x+c_{1}>0$ and $a_{2}x^{2}+b_{2}x+c_{2}>0$ are identical. Proposition $Q$: $\frac{a_{1}}{a_{2}}=\frac{b_{1}}{b_{2}}=\frac{c_{1}}{c_{2}}$. Then proposition $Q$ is a () condition for proposition $P$.\\ (A) Necessary and sufficient condition\\ (B) Sufficient but not necessary condition\\ (C) Necessary but not sufficient condition\\ (D) Neither necessary nor sufficient condition
\end{paperbox}

\begin{paperbox}{boxproblem}{Data sample: Physics (SciKnowEval)}
Question: A container with a volume of $18\,L$ contains a gas with a temperature of $270^{o}C$. If the temperature of the gas changes to $350\,^{o}K$ without any change in pressure, what must the container's new volume be?\\
A: 15.8 L \quad B: 11.6 L \quad C: 13.2 L \quad D: 9.4 L\\[2pt]
This is a multiple-choice question with exactly one correct option. Select the single best option and report ONLY its letter (A, B, C, or D) as your final answer — do not restate the option text.
\end{paperbox}

\begin{paperbox}{boxproblem}{Data sample: Material (SciKnowEval)}
Given the following crystal structure parameters for the material SmP$_{4}$HO$_{12}$ (Material ID: mp-542237), calculate the volume of the unit cell (in \AA$^{3}$).\\
Lattice Parameters:\\
a = 7.14 \AA \quad b = 8.58 \AA \quad c = 8.86 \AA\\
$\alpha$ = 99.11$^{\circ}$ \quad $\beta$ = 112.21$^{\circ}$ \quad $\gamma$ = 106.66$^{\circ}$\\
A: 458.96 \AA$^{3}$ \quad B: 465.00 \AA$^{3}$ \quad C: 482.10 \AA$^{3}$ \quad D: 473.25 \AA$^{3}$\\[2pt]
This is a multiple-choice question with exactly one correct option. Select the single best option and report ONLY its letter (A, B, C, or D) as your final answer — do not restate the option text.
\end{paperbox}

\begin{paperbox}{boxproblem}{Data sample: Tool-Use (ToolAlpaca)}
Your task is to answer the user's question using available tools.\\
You have access to the following tools:\\
Name: NetworkCalc\\
Description: Network calculators, including subnets, DNS, binary, and security tools\\
Documentation:\\
binaryToDecimal: Converts a binary number to its decimal equivalent.\\
Parameters: \{"binaryNumber": "Required. string. The binary number to convert to decimal."\}\\
Output: Decimal equivalent of the binary number.\\
\ \ - Format: application/json\\
\ \ - Structure: Object\{decimalNumber\}\\
\ \ [... further tools omitted for space ...]\\
Begin!\\
Question: I have a binary number, 101001110, and I need to know its decimal equivalent. Can you convert it for me?
\end{paperbox}

\section{Prompt Templates}
\label{app: prompts}

ReTeach uses two prompt roles that differ in \emph{how} the model is asked to reason, not in \emph{what} information it is given. The student answers in a single turn with thinking disabled, and therefore emits a solution followed by the final answer without any \texttt{<think>} block. The teacher answers with thinking enabled. When it is correcting a failed attempt, the prompt requires it to first write a structured reflection enclosed in \texttt{<reflect>} tags and then to solve the problem step by step, with the final answer in \texttt{\textbackslash boxed\{\}}. Because the teacher's native thinking is enabled, the model's internal \texttt{<think>} block appears before the externally required \texttt{<reflect>} block in the raw generation; the two are complementary, since the former is the model's own deliberation and the latter is the structured diagnosis we ask for. Neither role ever sees a reference solution. For the Math, Physics, and Material datasets, the outputs are straightforward to parse, so we use the generic student and teacher templates shown below.
\begin{paperbox}{boxprompt}{Student prompt (shared by Math, Physics, Material)}
\textbf{PROBLEM:} \textless{}question\textgreater{} \\ \\
Please reason step by step and put your final answer within \textbackslash{}boxed\{\}.
\end{paperbox}

\begin{paperbox}{boxprompt}{Teacher reflection prompt (shared by Math, Physics, Material)}
\textbf{PROBLEM:} \textless{}question\textgreater{}\\ \\ Below are previous incorrect attempts. Their final \textbackslash{}boxed\{\} answers have ALL been verified WRONG that you must avoid:\\
\textless{}Previous\_Incorrect\_Begin\textgreater{}\\ \textless{}Failed Attempts\textgreater{}\\ \textless{}Previous\_Incorrect\_End\textgreater{}\\ \\ \textbf{OUTPUT FORMAT:}\\ First, reflect on the previous attempts. Your reflection MUST BE enclosed within \textless{}reflect\textgreater{} \textless{}/reflect\textgreater{} tags, and inside it you must: name each attempt's failure type — repeated forbidden answer / wrong formula / calculation error / overlooked condition; re-derive the governing formula and re-check the numbers of the flawed step; state concretely how to correct it. After \textless{}/reflect\textgreater{}, solve the problem step by step and put your final answer within \textbackslash{}boxed\{\}. \\ \\
\textbf{IMPORTANT:} \\Your solution MUST be self-contained and stand entirely on its own. Write it as if you are solving the problem for the first time without any prior context.
\end{paperbox}

For the tool-use dataset, the model must be prompted in a task-specific manner~\citep{hubotter2026reinforcement} so that its output can be parsed as a ReAct-style transcript. We therefore provide the prompt templates below. They follow the same structure as those used for the other datasets, with the only addition being a specification of the required tool-call format.
\begin{paperbox}{boxprompt}{Student prompt (Tool-Use)}
\textbf{PROBLEM:} \textless{}question\textgreater{}\\ \\ Your final answer is the tool call itself. Inside \textbackslash{}boxed\{\}, write ONLY the tool call transcript, in exactly this format and nothing else:\\ Action: \textless{}one of the tool names listed above\textgreater{}\\ Action Input: \textless{}a single-line JSON object holding the arguments\textgreater{}\\ If the question needs several tool calls, put every Action / Action Input pair one after another inside that same \textbackslash{}boxed\{\}, in the order they should be executed.
\end{paperbox}

\begin{paperbox}{boxprompt}{Teacher reflection prompt (Tool-Use)}
\textbf{PROBLEM:} \textless{}question\textgreater{}\\ \\ Below are previous incorrect attempts. Their final \textbackslash{}boxed\{\} tool calls have ALL been verified WRONG — treat them as FORBIDDEN tool calls you must never produce again.\\ \textless{}Previous\_Incorrect\_Begin\textgreater{}\\ \textless{}Failed Attempts\textgreater{}\\ \textless{}Previous\_Incorrect\_End\textgreater{}\\ \\\textbf{OUTPUT FORMAT:}\\ First, reflect on the previous attempts. Your reflection MUST BE enclosed within \textless{}reflect\textgreater{} \textless{}/reflect\textgreater{} tags. An attempt is wrong if ANY of these holds — check them in this order: (a) MISSING or EXTRA call: the question often needs SEVERAL calls, and you must list them ALL in one transcript. (b) WRONG Action Input: every argument value must match the question CHARACTER FOR CHARACTER. (c) WRONG Action name: copy the tool name verbatim from the documentation, including its exact capitalization. After \textless{}/reflect\textgreater{}, solve the problem step by step and put your final answer within \textbackslash{}boxed\{\}. \\ \\
\textbf{IMPORTANT:} \\Your solution MUST be self-contained and stand entirely on its own. Write it as if you are solving the problem for the first time without any prior context.
\end{paperbox}

\paragraph{Teacher's reflection prompt is self-contained.} Although the teacher has access to the failed trajectory and the subsequent reflection, its prompt should only support the construction of a stronger response. We therefore require the teacher to produce a \uline{\emph{self-contained}} final solution and answer, following the same standalone response format as the student. In particular, the teacher must not refer to the failed history or information introduced during reflection. Otherwise, such context-dependent references would enter the distillation target and confuse the student.
\section{Generation Samples}
\label{app:generation}

We present a real training sample from the math dataset to show how ReTeach locates and corrects a reasoning mistake through reflect and retry. The example lists the inputs and outputs of the student and the teacher separately. The student receives the original question and produces an initial attempt that is verified as wrong. The teacher is then prompted with that attempt together with the verified-wrong answer. It first emits its native \texttt{<think>} block under thinking mode, followed by the \texttt{<reflect>} block fixed in the prompt template, the retried solution, and the final answer.

To facilitate comparison between the two responses, we highlight in \fcolorbox{red}{white}{red boxes} the incorrect part of the student's response and the part where the teacher identifies and corrects the error. The student enumerates only the factorizations in which both factors are equal, taking $4$ as $2 \times 2$ and $9$ as $3 \times 3$, and therefore omits the decompositions $4 = 1 \times 4$ and $9 = 1 \times 9$. Every valid pair involving $1$ is thus excluded and the resulting count is 3, which is verified as wrong. Within the \texttt{<reflect>} block, the teacher attributes the mistake to an incorrect count of valid pairs, particularly those involving 1, and then produces a correct derivation whose final answer matches the ground truth. Throughout the process, no external chain-of-thought or gold answer is introduced.

\begin{paperbox}{boxgen}{Generation sample: Student rollout and initial attempt on Math}
\textbf{PROBLEM:} Consider the cards $A, 2, \cdots, J, Q, K$ as the numbers $1, 2, \cdots, 11, 12, 13$. If we take the 13 cards of spades and 13 cards of hearts together and randomly draw 2 cards, what is the probability that the two cards are of the same suit and the product of the two numbers is a perfect square? \\ Please reason step by step and put your final answer within \textbackslash{}boxed\{\}.\\ \\
\textbf{STUDENT ATTEMPT:} \\
We are given 13 cards from spades and 13 cards from hearts, each numbered from 1 to 13 (i.e., representing numbers 1 through 13). So, in total, there are 26 cards.\\
We are to compute the probability that:
The two cards drawn are of the same suit (both spades or both hearts);
The product of the two numbers is a perfect square\\
Step 1: Total number of ways to draw 2 cards.\\
We choose 2 cards from the 26 cards:
$$
\text{Total outcomes} = \binom{26}{2} = \frac{26 \cdot 25}{2} = 325
$$\\
Step 2: Favorable outcomes\\
Let’s denote:
- $ S $: set of spade cards $\{1, 2, ..., 13\}$
- $ H $: set of heart cards $\{1, 2, ..., 13\}$
- So, we have 2 sets of 13 cards each, one for spades and one for hearts.\\
Case 1: Both cards are from spades\\
We need to count the number of pairs $(a, b)$ such that $a, b \in S$ and $ab$ is a perfect square.
Similarly for Case 2: Both cards are from hearts.
Since the set of numbers from 1 to 13 is the same for both suits, the number of such favorable pairs will be the same in both cases. So, we can compute the number of such pairs from one suit (say, spades), and then multiply by 2.\\
Step 3: How many pairs from a single suit (say, spades) have a product that's a perfect square?\\
We need to count the number of pairs $(a, b)$ where $a, b \in \{1, 2, ..., 13\}$, and $ab$ is a perfect square.
So, for each pair $(a, b)$ with $a \leq b$, we check if $ab$ is a perfect square.
Let’s list all possible pairs and identify those whose product is a perfect square.\\
Step 3.1: Check all pairs of integers from 1 to 13 and see which products are perfect squares.\\
Let’s consider all $1 \leq a < b \leq 13$, and check if $ab$ is a perfect square.
Note: We will only consider unordered pairs, since $(a, b)$ is the same as $(b, a)$ for the product.
Let’s list the perfect squares $leq$ 169 (since the maximum product is $13 \cdot 13 = 169$):
Perfect squares between 1 and 169 are:
$
1, 4, 9, 16, 25, 36, 49, 64, 81, 100, 121, 144, 169
$
So, we want pairs $(a, b)$ such that $ab$ is one of these values.
Now we search for all such pairs $(a, b)$ with $a < b$ and $ab \in \{1, 4, 9, 16, ..., 169\}$.
We'll compute all such pairs.
Let’s proceed with an approach:
For each perfect square $s$, find all factor pairs $(a, b)$ with $a < b$, $a \cdot b = s$, and $1 \leq a < b \leq 13$\\
Let’s proceed systematically:\\
$s = 1$: $a=1$, $b=1$ → discard since $a = b$
$s = 4$: $1 \cdot 4, 2 \cdot 2$ → discard $2 \cdot 2$ because $a = b$
$s = 9$: $1 \cdot 9, 3 \cdot 3$ → discard $3 \cdot 3$
$s = 16$: $1 \cdot 16$, $2 \cdot 8$, $4 \cdot 4$ → valid: $(1, 16)$, $(2, 8)$
$s = 25$: $1 \cdot 25$, $5 \cdot 5$ → discard $5 \cdot 5$
$s = 36$: $1 \cdot 36$, $2 \cdot 18$, $3 \cdot 12$, $4 \cdot 9$, $6 \cdot 6$ → valid: $(1, 36)$, $(2, 18)$, $(3, 12)$, $(4, 9)$
$s = 49$: $1 \cdot 49$, $7 \cdot 7$ → discard
$s = 64$: $1 \cdot 64$, $2 \cdot 32$, $4 \cdot 16$, $8 \cdot 8$ → valid: $(1, 64)$, $(2, 32)$, $(4, 16)$
$s = 81$: $1 \cdot 81$, $3 \cdot 27$, $9 \cdot 9$ → valid: $(1, 81)$, $(3, 27)$
$s = 100$: $1 \cdot 100$, $2 \cdot 50$, $4 \cdot 25$, $5 \cdot 20$, $10 \cdot 10$ → valid: $(1, 100)$, $(2, 50)$, $(4, 25)$, $(5, 20)$
$s = 121$: $1 \cdot 121$, $11 \cdot 11$ → discard
$s = 144$: $1 \cdot 144$, $2 \cdot 72$, $3 \cdot 48$, $4 \cdot 36$, $6 \cdot 24$, $8 \cdot 18$, $12 \cdot 12$ → valid: $(1, 144)$, $(2, 72)$, $(3, 48)$, $(4, 36)$, $(6, 24)$, $(8, 18)$
$s = 169$: $1 \cdot 169$, $13 \cdot 13$ → discard\\
Now, we eliminate any pair where $b > 13$, since the maximum number on the cards is 13.
So, for each of the valid pairs above, if $b \leq 13$, count the pair.
Go through them again and only include pairs where both $a$ and $b$ are in $[1, 13]$\\
Step 3.2: List of valid pairs where both numbers are in 1 to 13 and product is perfect square\\
Now let’s go back and find all pairs $(a, b)$, $a < b$, such that $a, b \in [1, 13]$, and $ab$ is a perfect square.
Let’s consider all perfect squares from $1^2 = 1$ to $13^2 = 169$, and for each one, list its factor pairs $(a,b)$ with $a < b$ and $a,b \leq 13$.
We’ll collect all such $(a,b)$ and count them.\\
\fcolorbox{red}{boxgen}{$s = 4$: $2 \cdot 2$ → $a = b$, so discard
$s = 9$: $3 \cdot 3$ → same, discard}
$s = 16$: $2 \cdot 8$, $1 \cdot 16$ → $16 > 13$ → only $(2,8)$
$s = 36$: $4 \cdot 9$, $3 \cdot 12$ → valid
$s = 64$: $4 \cdot 16$, $8 \cdot 8$ → $16 > 13$, 8 * 8 = same → only $(1, 64)$, invalid
$s = 81$: $3 \cdot 27$, $27 > 13$ → only $(1, 81)$
$s = 100$: $4 \cdot 25$, $5 \cdot 20$ → 25, $20 > 13$ → discard
$s = 121$: $11 \cdot 11$ → discard
$s = 144$: $6 \cdot 24$, $8 \cdot 18$, $4 \cdot 36$ → $all > 13$ → discard
$s = 169$: $13 \cdot 13$ → discard\\
\fcolorbox{red}{boxgen}{So the valid pairs are:
$(2, 8)$ → 16
$(3, 12)$ → 36
$(4, 9)$ → 36
That’s 3 valid pairs per suit.}\\
Step 4: Total favorable outcomes\\
Each suit has 3 such valid pairs → 2 suits → total favorable outcomes:
$
\text{Favorable} = 3 \text{ (spades)} + 3 \text{ (hearts)} = 6
$\\
Step 5: Compute the probability
$
\text{Probability} = \frac{\text{Favorable}}{\text{Total}} = \frac{6}{325}
$\\
Final Answer:
$
\boxed{\frac{6}{325}}
$ \\ \\
\textbf{VERIFIED:} Wrong\\[3pt]
\end{paperbox}

\begin{paperbox}{boxgen}{Generation sample: Teacher reflection and retry over student's failed attempts on Math}
\textbf{PROBLEM:} Consider the cards $A, 2, \cdots, J, Q, K$ as the numbers $1, 2, \cdots, 11, 12, 13$. If we take the 13 cards of spades and 13 cards of hearts together and randomly draw 2 cards, what is the probability that the two cards are of the same suit and the product of the two numbers is a perfect square?\\ Please reason step by step and put your final answer within \textbackslash{}boxed\{\}.\\ \\ 
Below are previous incorrect attempts. Their final \textbackslash{}boxed\{\} answers have ALL been verified WRONG that you must avoid:\\
\textless{}Previous\_Incorrect\_Begin\textgreater{}\\
(Student's reply above)$\cdots$
Final Answer:
$
\boxed{\frac{6}{325}}
$\\
\textless{}Previous\_Incorrect\_End\textgreater{}\\
\textbf{OUTPUT FORMAT:}\\ First, reflect on the previous attempts. Your reflection MUST BE enclosed within \textless{}reflect\textgreater{} \textless{}/reflect\textgreater{} tags, and inside it you must: name each attempt's failure type — repeated forbidden answer / wrong formula / calculation error / overlooked condition; re-derive the governing formula and re-check the numbers of the flawed step; state concretely how to correct it. After \textless{}/reflect\textgreater{}, solve the problem step by step and put your final answer within \textbackslash{}boxed\{\}. \\ \\
\textbf{IMPORTANT:} \\Your solution MUST be self-contained and stand entirely on its own. Write it as if you are solving the problem for the first time without any prior context.\\ 
\hrule height 0.4pt \hfill \\
\textbf{TEACHER REPLY:}\\
\textless{}think\textgreater{}\\
Okay, let's try to figure out this probability problem. The question is about drawing two cards from a combined deck of spades and hearts, each numbered 1 to 13. We need the probability that both cards are of the same suit and their product is a perfect square. The previous attempt gave 6/325, but that's forbidden. Let me see where they went wrong.\\
First, the total number of ways to draw two cards from 26 is indeed C(26,2) = 325. That part is correct. The problem is in counting the favorable outcomes. The previous attempt found 3 pairs per suit, leading to 6 total, but maybe they missed some pairs?\\
$\cdots$\\
Another possible mistake: perhaps they considered ordered pairs? But no, since they were considering unordered pairs. Or maybe they thought that (1,4) and (4,1) are different? But in combinations, they are same. So no. So the main error was missing those two pairs.\\
\textless{}/think\textgreater{}\\\\\\
\textless{}reflect\textgreater{}\\
The previous attempt calculated the total number of ways to draw two cards from 26 cards as $\binom{26}{2} = 325$, which is correct. It then proceeded to identify pairs of numbers from 1 to 13 that multiply to a perfect square. However, it made a critical error in counting the number of such pairs, leading to an incorrect result of $\frac{6}{325}$.\\
Key Mistakes:\\
1. **Missed Valid Pairs**: \fcolorbox{red}{boxgen}{The attempt only identified three valid pairs per suit}, but in reality, there are **five** such pairs per suit.\\
2. **Overlooked Condition**: The problem specifically requires the **same suit** and **product is a perfect square**. The previous attempt failed to account for all valid combinations, \fcolorbox{red}{boxgen}{particularly those involving 1 (e.g., $1 \times 4 = 4$, $1 \times 9 = 9$).}\\
3. **Calculation Error**: The count of favorable outcomes was \fcolorbox{red}{boxgen}{incorrectly limited to 3 per suit instead of the correct 5 per suit}, resulting in an incorrect total of 6 instead of 10.\\
\textless{}/reflect\textgreater{}\\\\
Step 1: Total Number of Outcomes\\
We are drawing 2 cards from a total of 26 cards (13 spades + 13 hearts):
$
\text{Total outcomes} = \binom{26}{2} = 325
$\\
Step 2: Count Favorable Outcomes\\
We need to find the number of unordered pairs $(a, b)$ such that:
$1 \leq a < b \leq 13$
$a \cdot b$ is a perfect square\\
We examine all possible pairs and identify those whose product is a perfect square. The perfect squares between 1 and $13 \cdot 13 = 169$ are:\\
$
1, 4, 9, 16, 25, 36, 49, 64, 81, 100, 121, 144, 169
$\\
Now, for each perfect square $s$, we find all pairs $(a, b)$ such that $a < b$, $a, b \in [1, 13]$, and $a \cdot b = s$. The valid pairs are:
$1 \cdot 4 = 4$
$1 \cdot 9 = 9$
$2 \cdot 8 = 16$
$3 \cdot 12 = 36$
$4 \cdot 9 = 36$\\
\fcolorbox{red}{boxgen}{So, there are **5 valid pairs** per suit.}\\
Since the cards are in two suits (spades and hearts), the total number of favorable outcomes is:\\
$$
\text{Favorable outcomes} = 5 \text{ (spades)} + 5 \text{ (hearts)} = 10
$$\\
Step 3: Compute the Probability\\
$$
\text{Probability} = \frac{\text{Favorable outcomes}}{\text{Total outcomes}} = \frac{10}{325}
$$\\
Simplifying the fraction:$\frac{10}{325} = \frac{2}{65}$\\
Final Answer:
$
\boxed{\frac{2}{65}}
$
\end{paperbox}
\paragraph{How the reflection is used at distillation time}
A teacher response consists of four parts: the native \texttt{<think>} block, the
required \texttt{<reflect>} diagnosis, the step-by-step solution, and the final answer. If the answer after retry is still wrong, only the
last two are carried into the next round as a failed attempt, so the history holds a solution and
its \texttt{\textbackslash boxed\{\}} answer. The distillation completion is always the student rollout $y$, as in Section~\ref{sec: ema}. During
distillation, ReTeach keeps the \texttt{<think>} and \texttt{<reflect>} blocks of the final
round only, strips them from the completion, and re-attaches them as a masked assistant prefix, so
the teacher is conditioned on its own deliberation when its logits are computed while the student
is trained on the solution and answer alone.

\end{document}